\documentclass[letterpaper, 10 pt, conference]{ieeeconf}  

\IEEEoverridecommandlockouts                              
\usepackage{textcomp}
\usepackage{gensymb}
\usepackage{graphicx}
\usepackage{subfig}
\usepackage{algorithm}
\usepackage{multirow}
\usepackage{multicol}
\usepackage[noend]{algpseudocode}

\usepackage{amsmath} 
\title{\LARGE \bf
Grid-based Localization Stack for Inspection Drones towards Automation of Large Scale Warehouse Systems
}

\author{ \parbox{6 in}
{\centering Ashwary Anand, Shubh Agrawal, Shivang Agrawal, Aman Chandra, Krishnakant Deshmukh \\
        Indian Institute of Technology (IIT) \\ 
        Kharagpur, West Bengal, India - 721302\\
 Email: {\{ash.anand, shubh.agrawal111, shivang1997agrawal, amanchandra333, krishdeshmukh24\}@iitkgp.ac.in}}
}

\begin{document}

\maketitle
\thispagestyle{empty}
\pagestyle{empty}

\begin{abstract}

SLAM based techniques are often adopted for solving the navigation problem for the drones in GPS denied environment. Despite the widespread success of these approaches, they have not yet been fully exploited for automation in a warehouse system due to expensive sensors and setup requirements. This paper focuses on the use of low-cost monocular camera-equipped drones for performing warehouse management tasks like inventory scanning and position update. The methods introduced are at par with the existing state of warehouse environment present today, that is, the existence of a grid network for the ground vehicles, hence eliminating any additional infrastructure requirement for drone deployment. As we lack scale information, that in itself forbids us to use any 3D techniques, we focus more towards optimizing standard image processing algorithms like the thick line detection and further developing it into a fast and robust grid localization framework.

In this paper, we show different line detection algorithms, their significance in grid localization and their limitations. We further extend our proposed implementation towards a real-time navigation stack for an actual warehouse inspection case scenario. Our line detection method using skeletonization and centroid strategy works considerably even with varying light conditions, line thicknesses, colours, orientations and partial occlusions. A simple yet effective Kalman Filter has been used for smoothening the $\rho$ and $\theta$ outputs of the two different line detection methods for better drone control while grid following. A generic strategy that handles the navigation of the drone on a grid for completion of the allotted task is also developed. Based on the simulation and real-life experiments, the final developments on the drone localization and navigation in a structured environment are discussed.

\end{abstract}

\section{INTRODUCTION}

Use of drones has boomed in the past few years especially towards the industrial usage that ranges from delivery to fault checks. All these applications have a standard fundamental requirement i.e., smooth navigation of the robot from an initial location to the final location, for which the robot needs to know it's local/global position in real-time. Indoor localization introduces an another layer of difficulty as the presence of GPS signal is generally weak or unavailable for the drones to localize in a global frame. This type of scenario is commonly found in industrial and warehouse environments that more or less functions in the indoor conditions. Figure 1 shows a scenario from a large warehouse system updating towards its automation.

For any mobile robot working in an application, the primary task it must be excelling at is navigation. There are various navigation strategies used for mobile robots depending upon environment viz. indoor, outdoor and depending upon information viz. map-based, map-less. Also, just designing an algorithm is not sufficient. For navigation, various sensors are assembled onboard the mobile robot that senses the surrounding environment and acts as a feedback. This feedback ensures that the robot follows the commands given by the algorithm as accurately as possible. Today, various sensors like LiDAR, Camera, Radar, Sonar, etc., are explicitly designed for onboard use on mobile robots. However, researchers now are focusing more towards camera-based navigation as it is substantially cost-effective compared to the rest (Radar, LiDAR are still expensive) and provides detailed information about the environment which may not be available even after using a combination of the other sensors. UAVs perform active manoeuvres in their application environment which limits their size and further reducing their payload carrying capacity. Industrial applications that demand significant UAV flight time should also be light and cost-efficient. Thus, efforts are made to reduce the number of onboard sensors. For all these requirements, a smart usage of the  monocular camera can serve the overall purpose.

\begin{figure}[ht!]
  \centering
  {{\includegraphics[width=8.5cm]{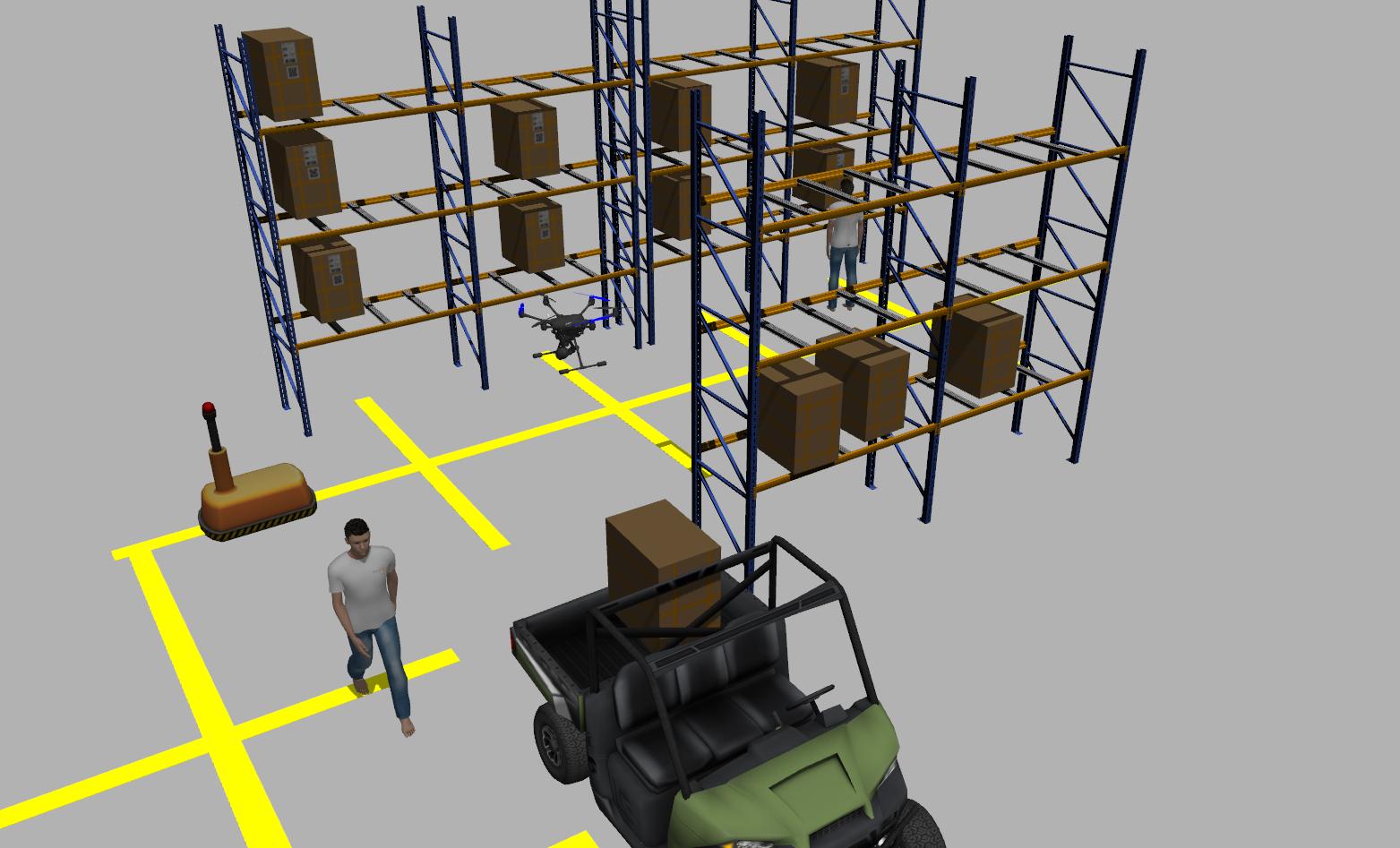}}}
 \caption{Cropped view of a typical large scale warehouse environment automated with the help of ground robots; The presented approach suggests the use of existing guidelines for the deployment of inspection drones}
\end{figure}

\begin{figure}[ht!]
  \centering
  {{\includegraphics[width=8.5cm]{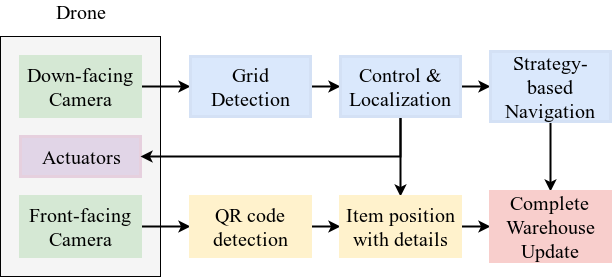}}}
 \caption{Overview of the proposed workflow for warehouse inventory management by the drone}
\end{figure}

A navigation stack will only make sense if the drone is capable of localizing in an environment. For the presented case, we need to localize using the lines and nodes existing in the environment for which we need to detect the thick guidelines first. One of the most widely used line detection algorithms is Hough-line-based detection, but it has many limitations that are expected to create problems in a real-world scenario. Much of the efficiency of the Hough transform \cite{Hough_1962} relies upon the quality of input data i.e., the edges have to be detected well for the Hough transform to be efficient. Using the vanilla hough transform on the noisy images is considerably prone to the haywire detections, and generally, to tackle this, a denoising stage is prefixed. A rule-based denoising framework leads to the occurrence of variance by threshold configurations, whereas, a learning-based denoising framework leads to an increase in computational load. Hence, we propose a pipeline for robust line detection that overcomes these problems by reducing the artifacts to a state easily detected by hough line algorithm with less severe errors.

A standard thick line detection algorithm generally supports the following steps:
\begin{enumerate}
\item Denoising
\item Line Segmentation / Thresholding 
\item Edge Detection
\item Hough Line Transform
\item Combining multiple results
\end{enumerate}

The pipeline we propose in this paper, as shown in Figure 2, starts with subscribing to the down-facing camera images from the drone platform. A thick-line detection algorithm was implemented as a part of machine vision which holds significant value for the optimal performance of localization. The paper is divided into following sections. Literature review and relevant studies in this direction are discussed in Section II. Sections III states all the assumptions and conventions considered while the development of the framework. Section IV gives a walk-through of different methods applied with a brief discussion. Section V describes the control and localization implementation for the drone using the output from our thick-line detection algorithm. Any offset of drone from the closest safe point of hovering (i.e. intersection of two guidelines) is reduced with multiple PID loops acting on the lines detected within the image. Thus, discretizing the drone's location to nodes co-ordinates. Finally, section VI discusses a navigation strategy that we suggest for the drone to follow in the given structured environment.

\section{RELATED WORK}
Large warehouses across the country lose millions of dollar every year due to lost inventory. Manual scanning of inventory is both time-consuming and prone to error. This problem of real-time warehouse inspection system has been approached by different entities. Kiva Systems, now owned by Amazon, has implemented an autonomous inventory system in their warehouses that track both the location and contents of every bin in the warehouse using AGVs. Few of the problems that this system face include a limited height of the shelf, a requirement of a manual process for the inventory scanning and low-speed manoeuvre of the ground vehicles. Another option available in the industry involves deploying an infrastructure of permanent shelves with built-in RFID readers such that the shelf can recognize the item being stored in it. While this system gives a real-time scan of the inventory, the permanent shelf doesn't allow the ground robots to automate the inventory movement.

MIT's RFly \cite{rfly} used relays which during the flight picked RFID responses from various sites along the path of the drone, and handled these spatial measurements as an antenna array. Later, by applying the antenna array equations to these measurements, they can localized RFIDs. This method is based on previous work on Synthetic Aperture Radar (SAR) systems. Another organization, Infinium Robotics, used a miniature ground vehicle mounted with a visual tag for the drone to follow between the shelf gaps. This type of method increases the cost per drone because of the requirement of an additional ground robot to act as a guide.

For all such drones developed towards warehouse automation, localization stands as the most critical module. Hence, we considered it important to review the present state of localization methods for structured environments. The current state-of-the-art algorithms in this direction are majorly the existing algorithms in simultaneous localization and mapping (SLAM) for autonomous robots using different sensor setups. For example, LSD-SLAM \cite{lsdstereo} and ORB-SLAM2 \cite{orb} used stereo camera, DSO \cite{dso} and VINS-Mono \cite{vins} used monocular camera with inertial sensor, the work presented by Hess et. al. \cite{2dlidar} and Bosse et. al. \cite{bosse} used LiDARs, etc. Although these methods work quite well even in unstructured environments, sensors like stereo camera and LiDAR systems are, in general, bulky to be used on-board UAVs in a constrained environment with a considerable cost of deployment for large-scale application. Moreover, these systems have high computational power and memory need that renders them unsuitable for on-board use of a micro-aerial vehicle (MAV). Our method, therefore, utilizes the presence of structured environment, like the use of grid lines available in typical warehouses for the movement of ground robots, to build a robust and computationally efficient solution that allows the drone to localize itself on a warehouse map with a sufficient accuracy. A previous grid-based localization system \cite{manash} used the properties of particularly structured grid-lines to localize a MAV inside each sub-cell of the grid. However, it required a minimum of three vertical and horizontal equally spaced grid lines for sufficiently accurate localization. Our method removes this extra constraint of having multiple lines by employing a combination of elegant and simpler strategies with discrete localization on nodes.

\section{ASSUMPTIONS AND CONVENTIONS}
\subsection{Conventions}

The following conventions are used in this paper: 
\begin{enumerate}
\item The drone's coordinate system: X-axis and Y-axis are along the roll and pitch axis of the drone in hovering mode respectively, the origin being the centre of the drone.
\item Vertical line: The straight line parallel to the shelf.
\item Horizontal line: The straight lines perpendicular to the shelf.
\item $\theta$ : Angle made by the perpendicular to the vertical line from origin with the X-axis in drone's co-ordinate system
\item  $\rho$ : The length of the perpendicular to the vertical line from origin with the X-axis in drone's Co-ordinate system. Figure 3 shows a sample snapshot containing drone with its co-ordinate axes, the vertical line, the parameters $rho$ and $theta$.
\item The pitch angle of the drone is along the horizontal line and roll angle along the vertical line.
\item -10$\degree$ $\leq$ roll $\leq$ 10$\degree$, though we only used to give roll in range of -1$\degree$ and 1$\degree$.
\item -10$\degree$ $\leq$ pitch $\leq$ 10$\degree$, though we only used to give pitch in range of -1$\degree$ and 1$\degree$.
\item -10$\degree$ $\leq$ yaw-rate $\leq$ 10\degree,  though we only used to give yaw-rate in range of -1$\degree$ and 1\degree.
\end{enumerate}

\begin{figure}[ht!]
  \centering
  {{\includegraphics[width=8cm]{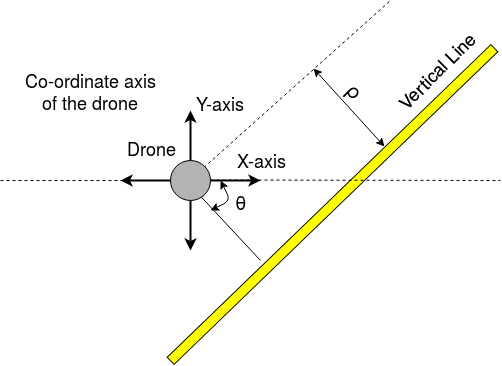}}}
 \caption{$rho$ ($\rho$) and $theta$ ($\theta$) convention with respect to drone's local co-ordinate axis}
\end{figure}
 
\subsection{Assumptions}
A warehouse is a well-structured environment. The boxes or cargo placed in a warehouse shelf may have different sizes i.e., there may not be a specific constant horizontal distance between any two adjacent boxes. To take such situations into account and to ensure that no box is left unscanned, some acceptable conditions have been assumed. The lines and nodes are assumed to be next to the cargo shelf where drone needs to stop and scan for the QR codes and report the location corresponding to the nodes. A typical warehouse setup following the stated assumptions is shown in Figure 4. The horizontal line should end before some distance ($D_Y$) from the shelf to avoid any contact of the drone with the boxes. $D_Y$, $D_X$ (distance between any two adjacent nodes) and $\sigma$ can be related by a simple equation given by   
\begin{equation}
D_X = 2\times D_Y \times\tan( \frac{\sigma}{2} )
\end{equation}
where 

\qquad $\sigma$ $\gets$ Field of view of the mounted camera. Equation (1) can be easily realized from Figure 2.

The vertical length between the start of any shelf and the first node must be less than or equal to $\frac{D_x}{2}$. 
\begin{figure}[ht!]
  \centering
{{\includegraphics[width=8cm]{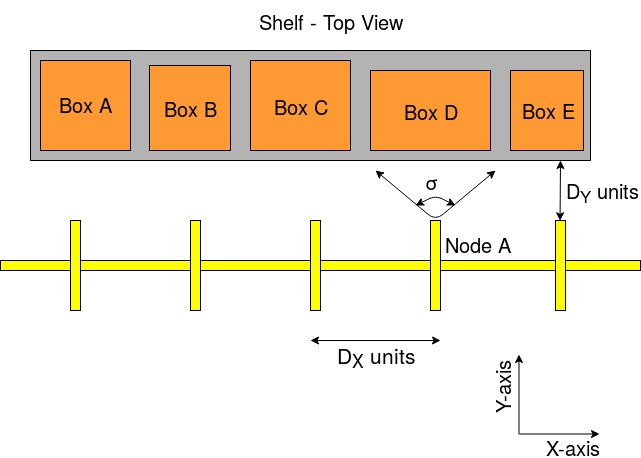}}}
  \caption{A typical Warehouse setup}
\end{figure}

For a controlled environment like a warehouse, constant pressure with negligible wind is assumed to minimize the errors in height estimation using a barometer, good lighting condition is expected for crisp grid detection and QR code detection. Any moving obstacle like forklifts, overhead chains, etc which can hinder the path of the drone is assumed to be absent or away from the drone's route.

\section{VISION SYSTEM}
For the presented work, vision algorithms hold a crucial position for the drone navigation as it allows the drone to localize within the grid structure. In such a scenario, using a simple method consisting of canny and hough transform as line detection method might seem promising, but is equally risky to use citing its naivety under extreme conditions. Hence, it is necessary to improve certain sections of the grid detection algorithm focusing towards robustness while also keeping the time complexity similar to the simple method as above. Following sub-sections discuss some of the improvements, we propose.

\subsection{Line Thresholding}
Image input is converted to HSV colour space for thresholding as it provides a better thresholding accuracy compared to RGB colour space.
A Support Vector Machine model \cite{svm} was used to train the thresholding of different colour segments and used region growing algorithm on top of it to identify the line bands for varying brightness and intensities. We trained our model to detect yellow colour under varying conditions and lighting. Results were tested on blurred and distorted video and our model was robust enough to quickly threshold the line with minimal loss in accuracy. Figure 5 shows the effectiveness of colour segmentation using the above method. 

\begin{figure}[ht!]
  \centering
  \subfloat[Input Camera Feed]{{\includegraphics[width=3.7cm]{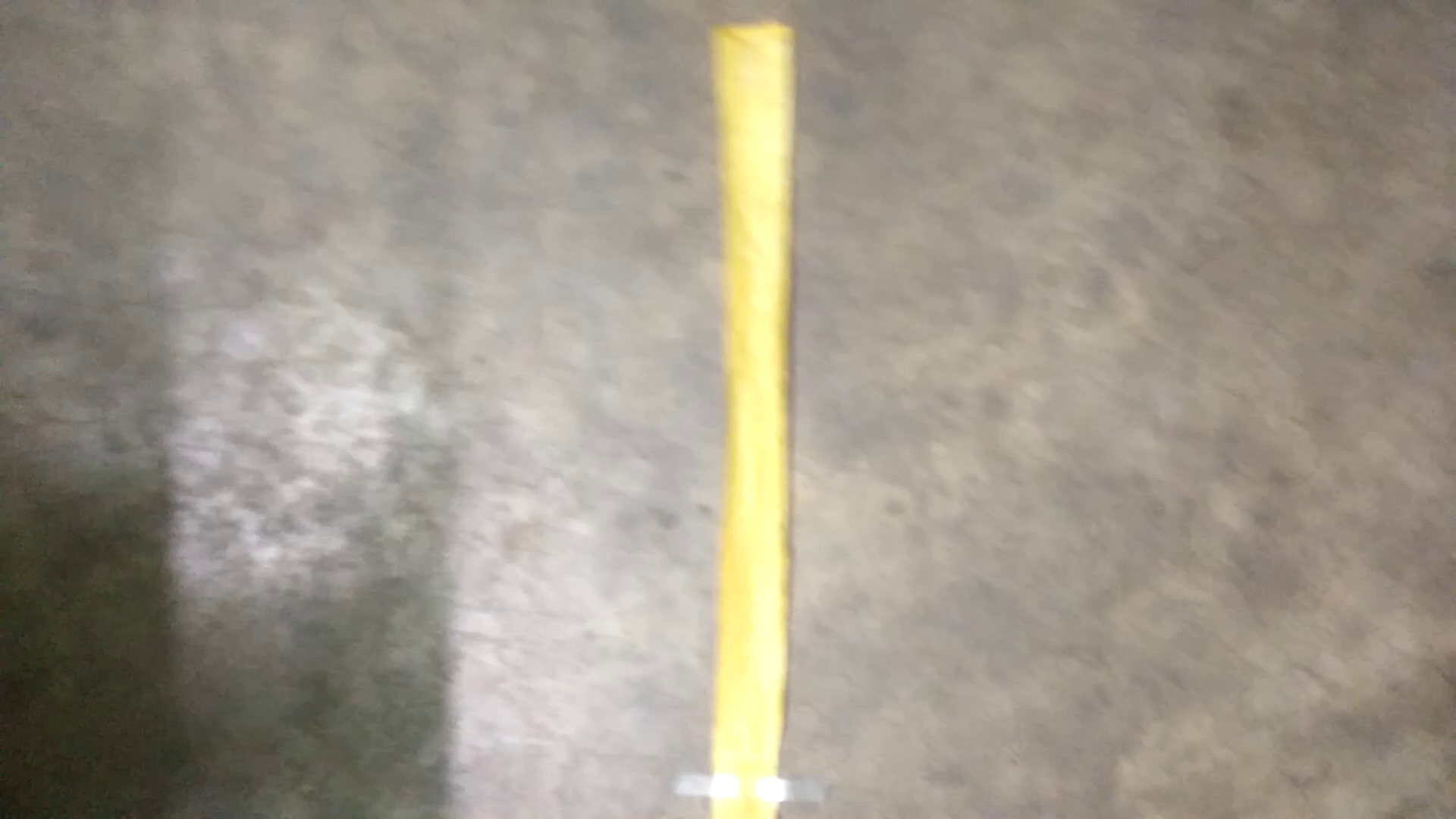}}}
  \qquad
  \subfloat[Thresholded Line]{{\includegraphics[width=3.7cm]{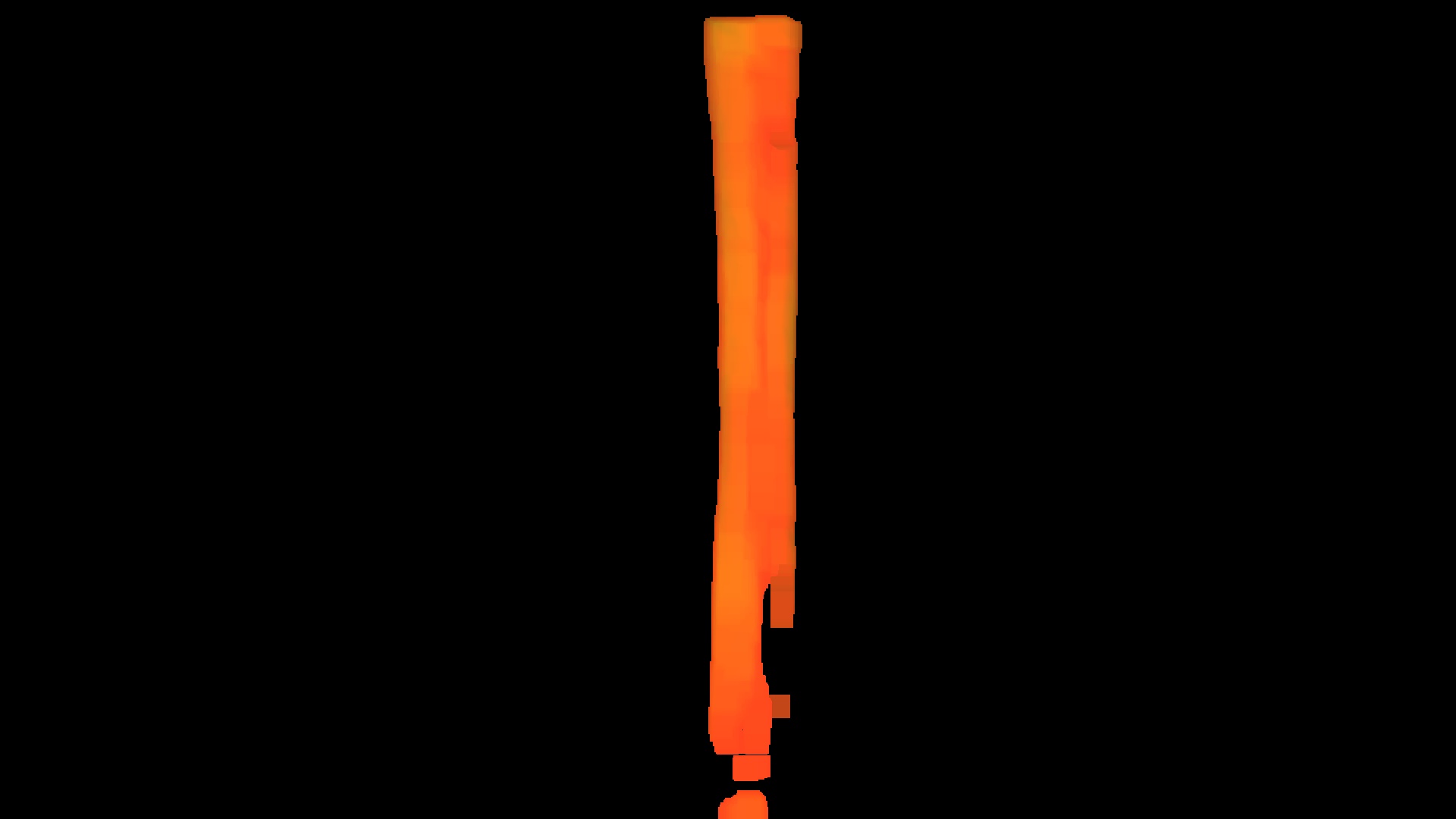}}}
  \caption{Line Thresholding in HSV Space using SVM}
\end{figure}

\subsection{Line Tracking}
\subsubsection{Centroid Method}
The thresholded image was broken into $W\times H$ slices which are dynamically identified based on the lines coverage area in the image, e.g., increasing the altitude of drone reduces the thickness of line thus greater slices are needed to detect the line features properly. For a very large area of lines in the image, the slices can be reduced, while for a small area of lines more slices are needed to cover all the curves and turn in the lines. Each slice was processed in parallel to find the centroid of each slice using OpenCV and contour detection. Moments of each contour was calculated and the centroid was returned for each slice of the image. Figure 6 shows the centroid extraction result on a sample image.\\
Moment of any object in 2D space is given by
\begin{equation}
	M_{pq} = \int_{-\infty}^{\infty}\int_{-\infty}^{\infty}x^py^qf(x,y)\mathrm{d}x\mathrm{d}y
\end{equation} 
For a grayscale image with pixel intensity I(x,y) , raw moment is given by 
\begin{equation}
	M_{ij} = \sum_x\sum_yx^iy^jI(x,y)
\end{equation}

Centroid of the object then can be calculated as 
\begin{eqnarray}
	\bar{x} = \frac{M_{10}}{M_{00}}\\
    \bar{y} = \frac{M_{01}}{M_{00}}
\end{eqnarray}
Image slices were then concatenated to form an image with the centroid points. These centroid points were used for RANSAC \cite{ransac} model fitting to find the best fitting line for the given set of points.

\begin{algorithm}
\caption{Centroid Extraction}\label{alg:centroid}
\begin{algorithmic}[1]
\Procedure{Centroid}{$image,N_{height},N_{width}$}
\State $images\gets divide(image,N_{height},N_{width})$
\For{\texttt{img \textbf{in} images}}
    \State $contours\gets contour(img)$
	\State $contour_{Max} \gets Max_{area}(contour)$
	\State $M \gets moment(contour_{Max})$
	\State $Centre_X = M_{10}/M_{00} $
	\State $Centre_Y = M_{01}/M_{00}$
	\State \textbf{push} $Centre_X, Centre_Y$
\EndFor
\State $image \gets concatenate(images)$
\State \textbf{return} $image$
\EndProcedure
\end{algorithmic}
\end{algorithm}

\begin{figure}[ht!]
  \centering
  \subfloat[Thresholded Line]{{\includegraphics[width=3.7cm]{Pic.jpg}}}
  \qquad
  \subfloat[Centroid separation]{{\includegraphics[width=3.7cm]{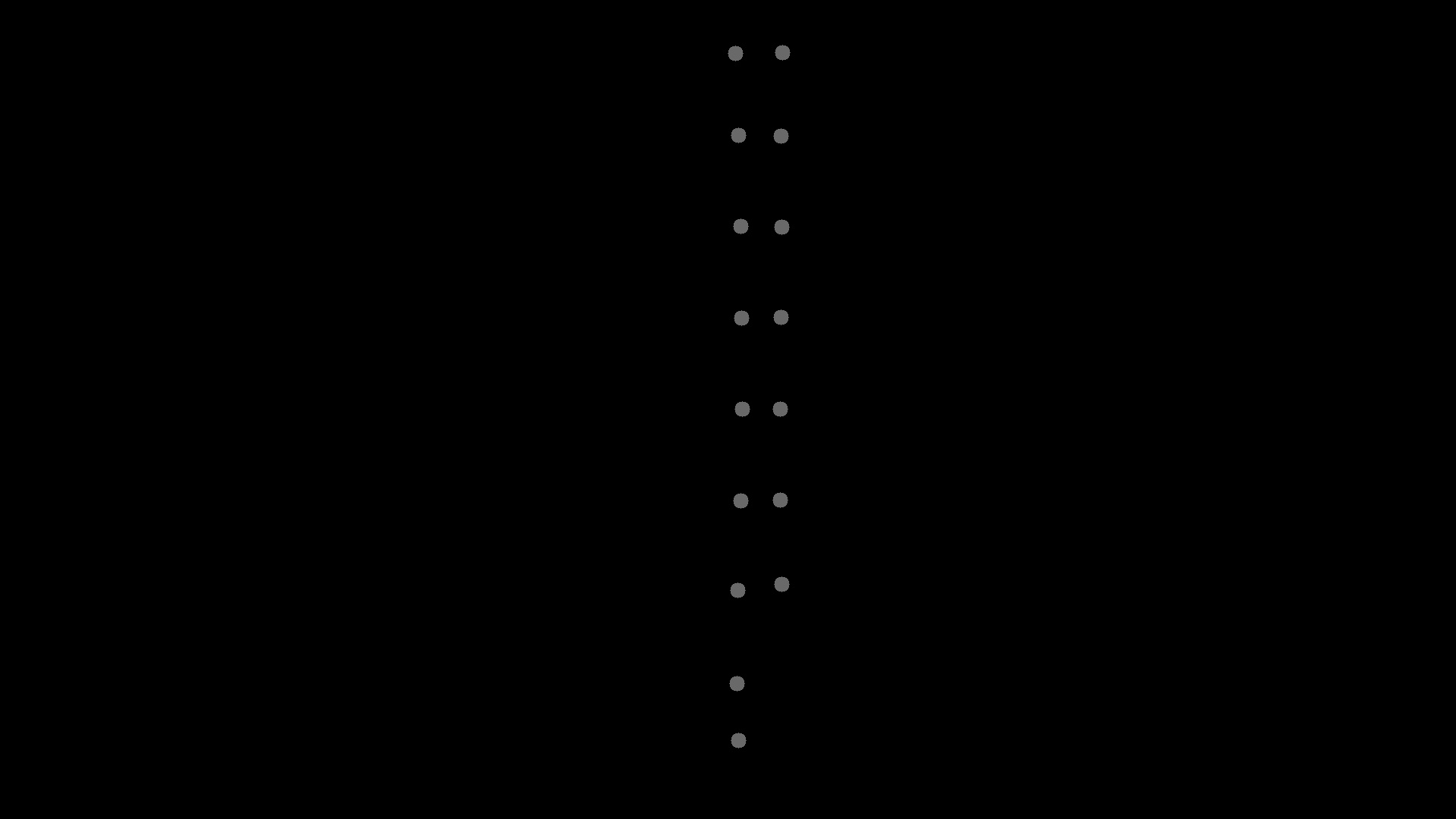}}}
  \qquad
  \subfloat[Centroid of sub-parts]{{\includegraphics[width=3.7cm]{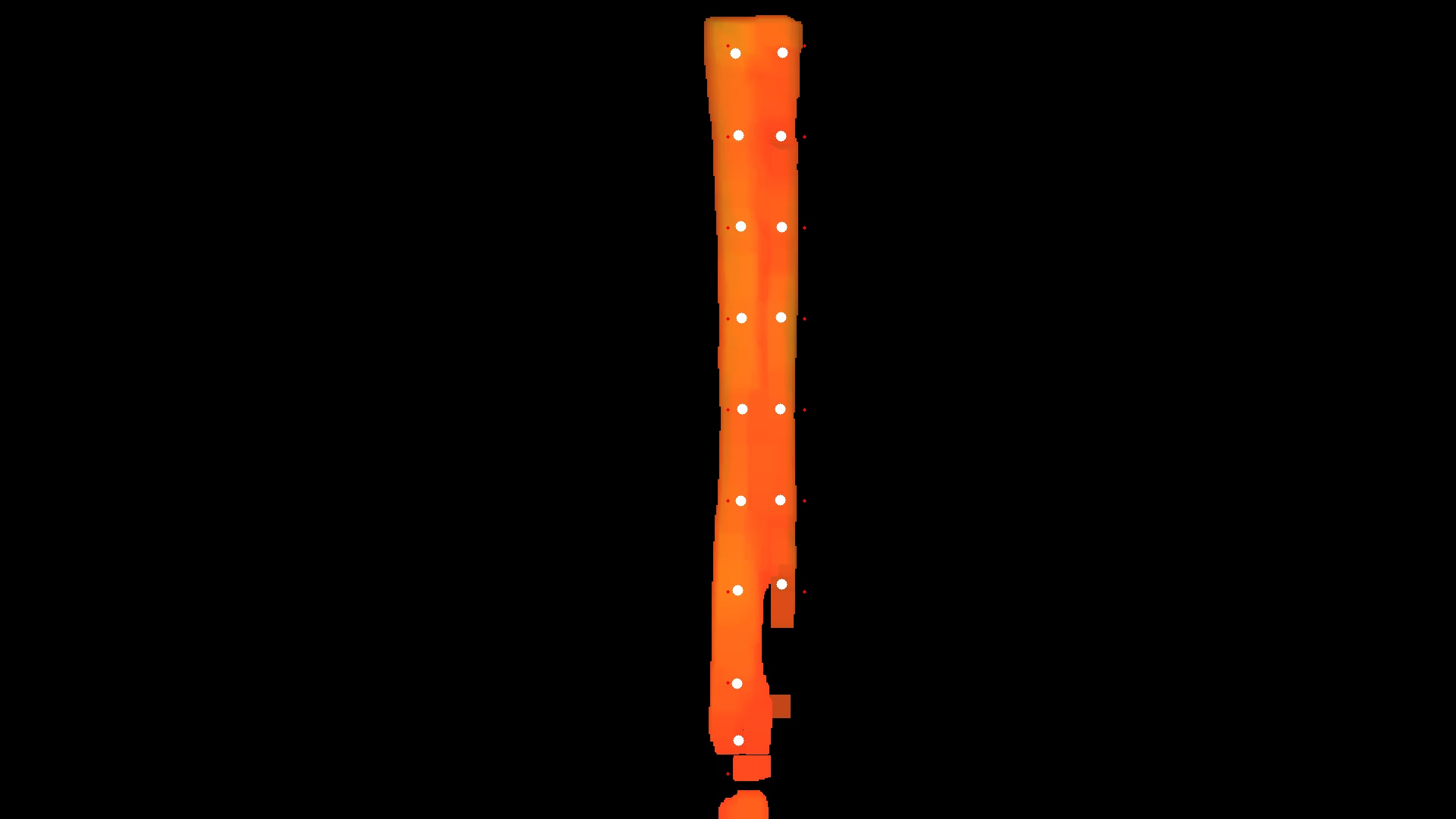}}}
  \caption{Centroid Detection}
\end{figure}

\subsubsection{Skeletonization}
The centroid method can get computationally expensive for an on-board processor fitted on drones like Raspberry Pi; an alternative method can be used to attain respectable performance compared to the centroid method. 

The morphological skeleton of a shape is a thin version of the form that is formed at the centre of the shape, equidistant from the edges. It usually emphasizes the geometric and topological features of the structure, such as shape, topology, length, direction, and width.

\begin{figure}[ht!]
  \centering
  \subfloat[Input Camera Feed]{{\includegraphics[width=3.7cm]{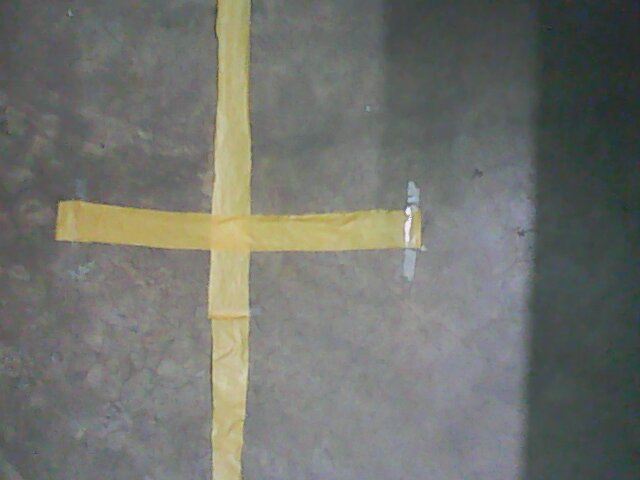}}}
    \quad
    \subfloat[Canny Edge Detection]{{\includegraphics[width=3.7cm]{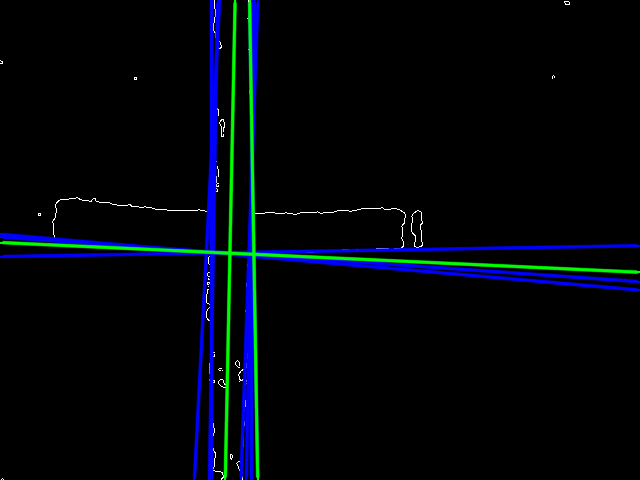}}}
    \quad
    \subfloat[Skeleton Extraction]{{\includegraphics[width=3.7cm]{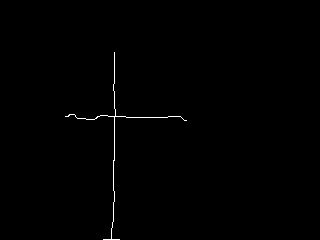}}}
    \quad
    \subfloat[Ours]{{\includegraphics[width=3.7cm]{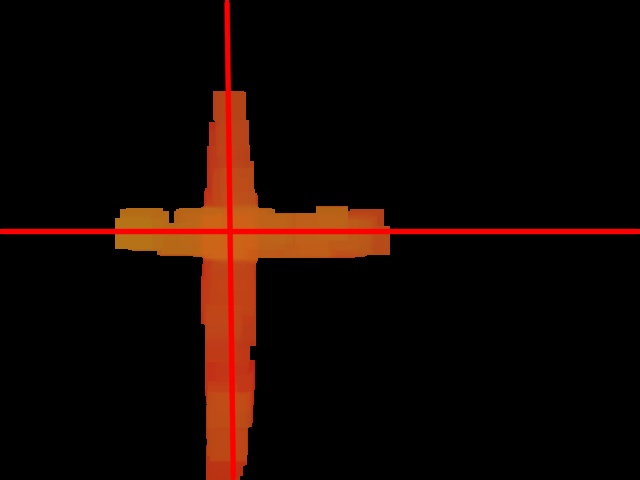}}}
  \caption{Line detection comparison between the use of skeleton extraction and canny edge detection before the application of hough line transform}
\end{figure}
 
The thresholded image was converted to binary and image was eroded to obtain the skeleton view of the image. Thinning algorithm by Zhang-Suen \cite{thinning} was used to thin the contours to a few pixels wide producing a single straight line for the hough line to work properly. Hough transform was then used to find the lines from the reduced skeletonized points. Figure 7 shows one such sample of application of this method with comparison with canny edge detection. To further reduce the error, mean clustering was used to cluster lines to an accuracy of $5$ degrees. Two ranges of $\theta$ were considered for classifying as a vertical or horizontal line.
 
\begin{equation}
	Line = \left\{
  			\begin{array}{@{}ll@{}}
    			vertical, & \text{if}\ 0\degree \leq  \theta < 30\degree  \\
    			vertical, & \text{if}\ 150\degree <  \theta < 180\degree  \\
                horizontal, & \text{otherwise}  \\
  			\end{array}\right.
\end{equation} 
 
 \begin{figure}[ht!]
  \centering
  {{\includegraphics[width=7cm]{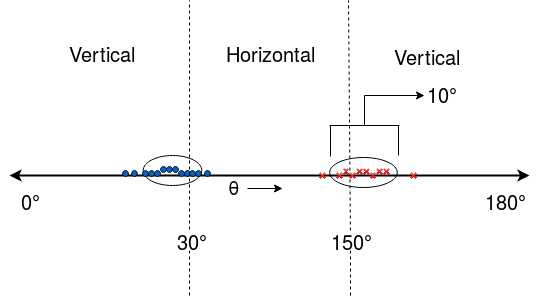}}}
  \caption{Clustering of Lines based on $\theta$}
\end{figure}
 
Lines are sorted into vertical and horizontal topics using means clustering with a specific threshold and are clustered into a group of two lines. Figure 8 shows the threshold range and permitted perturbation for the line clustering.

If the skeletonization method is used for line detection, there may be an offset in the line detected because of the nature of skeletonization algorithm. This can be tackled by fusing the data from centroid as well as skeletonization, if the hardware permits, to gain more accurate line detections.  

\section{CONTROL AND LOCALISATION}

The centroid or skeletonization algorithm returns a set of lines that sometimes are found to be fluctuating by a small quantity, which then affects the stability of the drone while following the line. Therefore, a simple Kalman filter was applied on the output line equations (parameters in the Hessian form) to minimize the effect of jitters and random false lines produced by the cluttered noise in the image.

A clustering algorithm was also implemented in case of skeletonization which further normalizes any erroneous data produced due to the branching of the skeleton of the line i.e., multiple perpendicular lines to the reference line occurring in an image. Assuming resolution of the line to be in the range of $5\degree$, lines were clustered in different pairs with $5\degree$ difference between them. Figure 9 shows the detection of two nodes after applying this method to a sample image. This helps to provide yaw feedback to the drone with respect to the line and can correct any other orientation errors induced in the system.
\begin{figure}[ht!]
  \centering
  \subfloat[Input image with 2 nodes]{{\includegraphics[width=2.8cm]{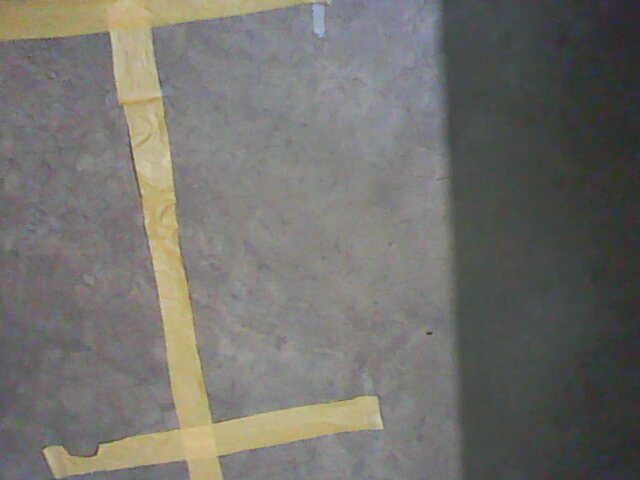}}}
  \subfloat[Averaging effect]{{\includegraphics[width=2.8cm]{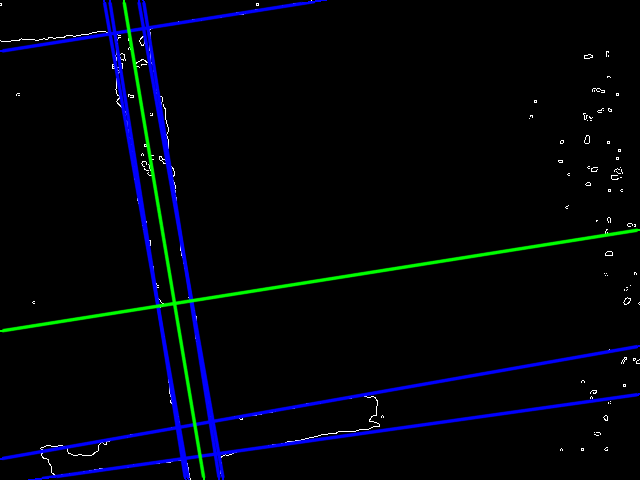}}}
  \subfloat[Clustering rho, theta]{{\includegraphics[width=2.8cm]{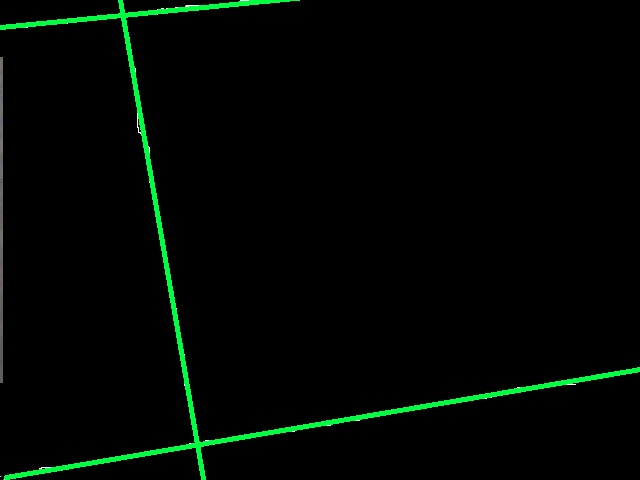}}}
  \caption{Effectiveness of the clustering algorithm on multiple line detections for identifying the next target or the closest node}
\end{figure}

 A closed-loop PID controller was used on the roll, pitch and yaw rate of the drone with the lines pose as feedback to control the $x$ and $y$ position, and yaw of the drone with respect to the grid lines. The design of PID controller that was implemented for our purpose is shown in Figure 10. The velocity output of each of the control loop was clipped to a range $[-0.1, 0.1]$ to avoid the possibility of drone deviating by unexpectedly high value and the grid line getting out of the field-of-view of the down-facing camera.
PID equation of each subsystem is given as follows:

\begin{equation}
PID = K_P e(t) + K_I\int e(t) + K_D\frac{de(t)}{dt}
\end{equation}

where:

\qquad $K_P$: Proportional Constant

\qquad $K_I$: Integral Constant

\qquad $K_D$: Derivative Constant

\qquad $w$: Image Width

\qquad $h$: Image Height

\qquad $\Delta x$ = $\frac{w}{2}$ - $\rho_{vertical}$

\qquad $\Delta y$ = $\frac{h}{2}$ - $\rho_{horizontal}$

\qquad $\Delta \theta$ = 0 - $\theta_{horizontal}$

\qquad integral = $integral_{previous}$ + error

\qquad derr = error - $error_{previous}$

\qquad dt : execution time

\begin{figure}[ht!]
  \centering
    {{\includegraphics[width=8cm]{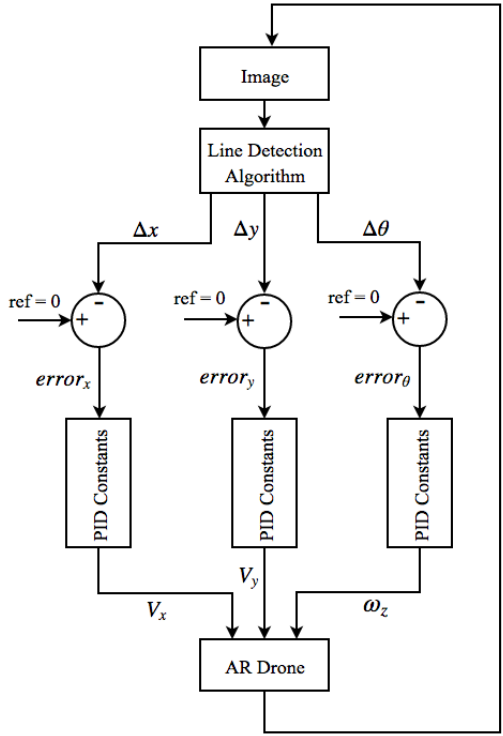}}}
  \caption{Block diagram of our implemented control system.}
  \centering
\end{figure} 

Even though the roll and pitch angles were clipped to a small value, the instantaneous orientation of the drone created an offset in the position of the line in an image as shown in Figure 11. This offset was observed to be proportional to the altitude (H) and tangent (trigonometric function) of the roll or pitch angle of the drone for vertical line and horizontal line respectively. To compensate for this offset for accurate estimation of the position of lines, we implemented an offset correction step before sending the line error signal to the controller.
The correction step is formulated as follow:

\begin{equation}
\rho_{vertical}^{Corrected} = \rho_{vertical} - H\times\tan(roll)
\end{equation}
\begin{equation}
\rho_{horizontal}^{Corrected} = \rho_{horizontal} - H\times\tan(pitch)
\end{equation}

Finally, the PID control loops were tuned independently one by one first and then together using the previously estimated parameters as the starting point. Figure 14 shows the experimental setup along with the drone used for tuning the parameters.

\begin{figure}[ht!]
  \centering
  \subfloat[0\% of maximum roll.]{{\includegraphics[width=2.8cm]{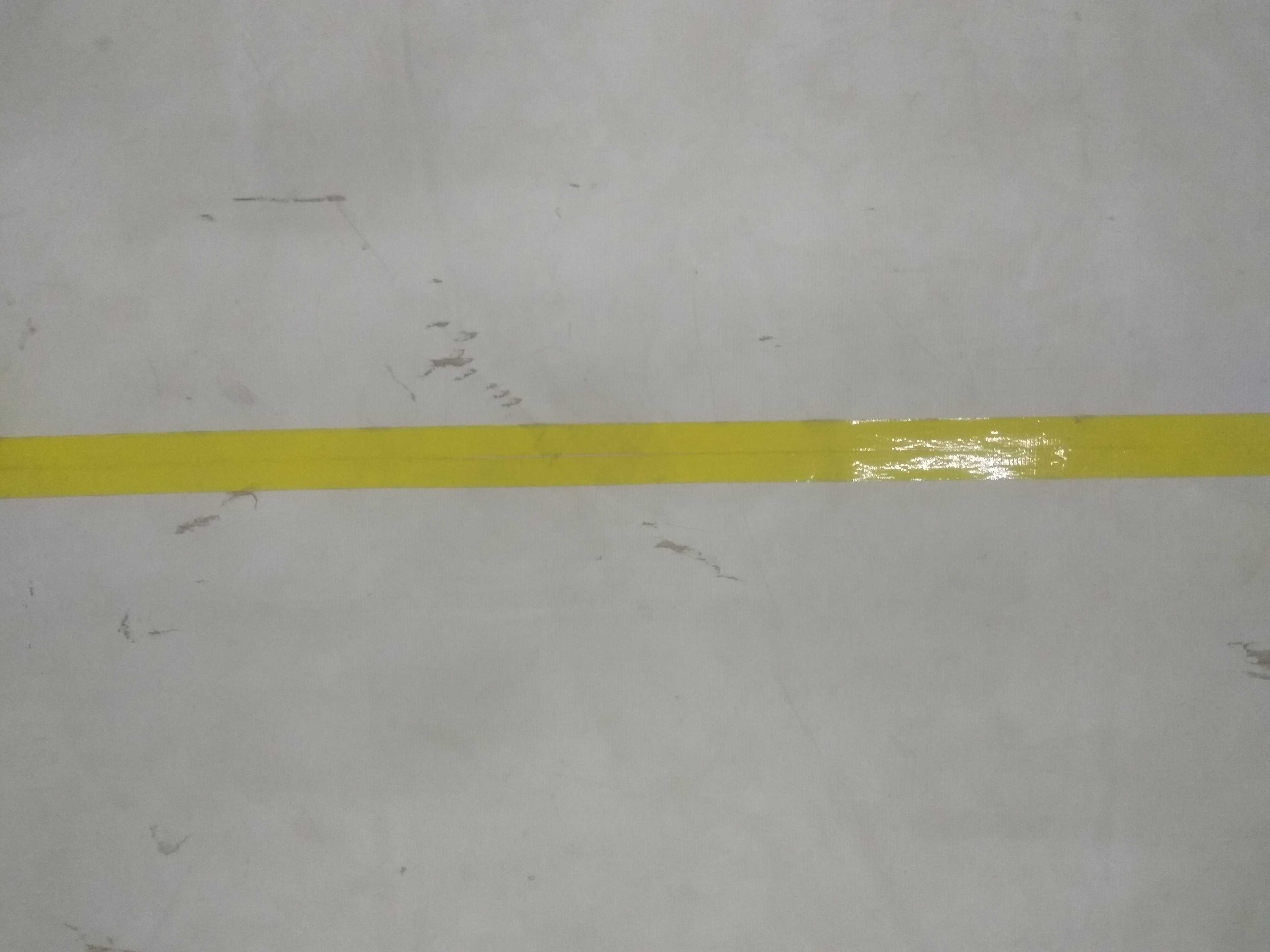}}}
  \subfloat[50\% of maximum roll.]{{\includegraphics[width=2.8cm]{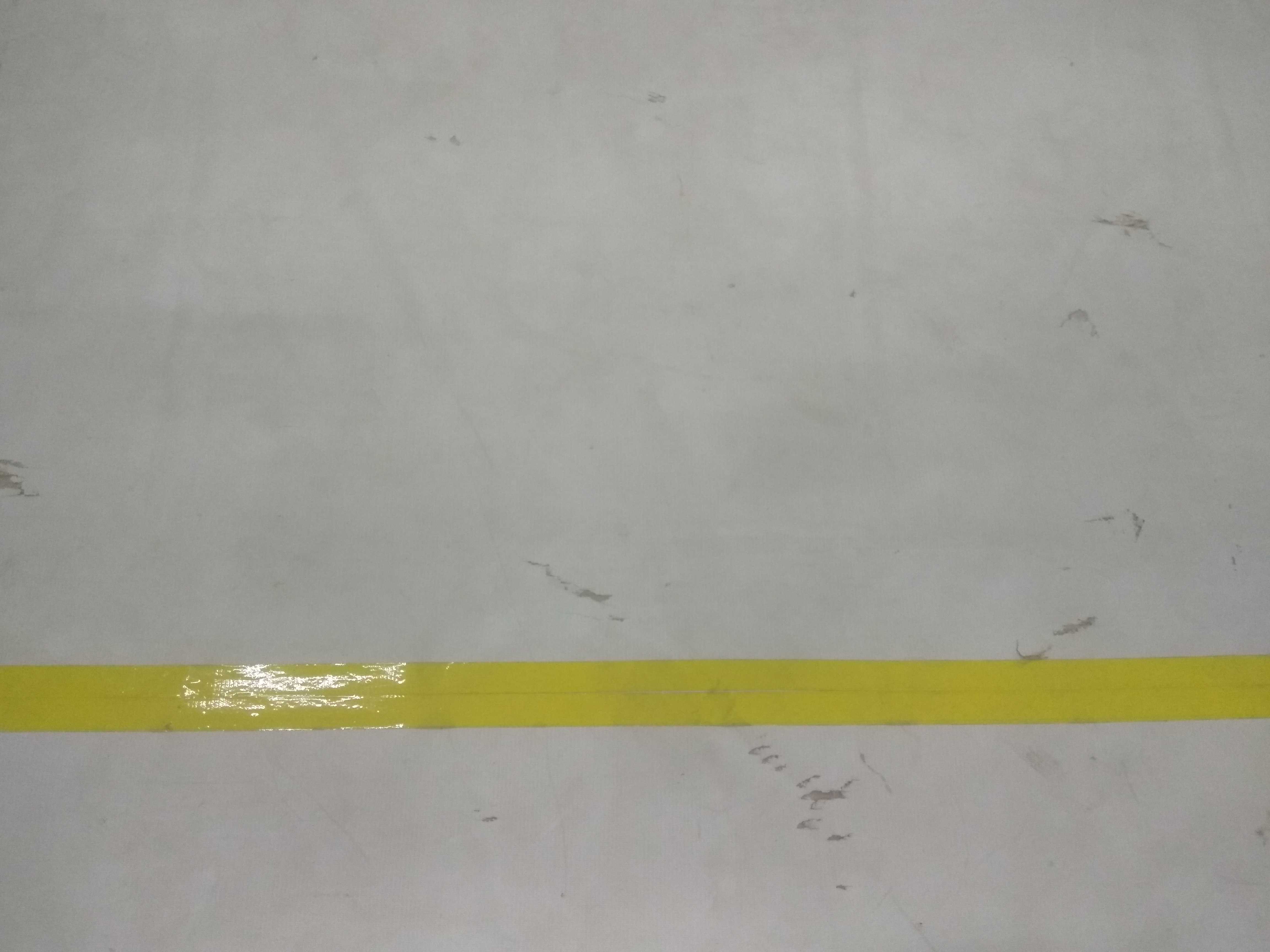}}}
  \subfloat[90\% of maximum roll.]{{\includegraphics[width=2.8cm]{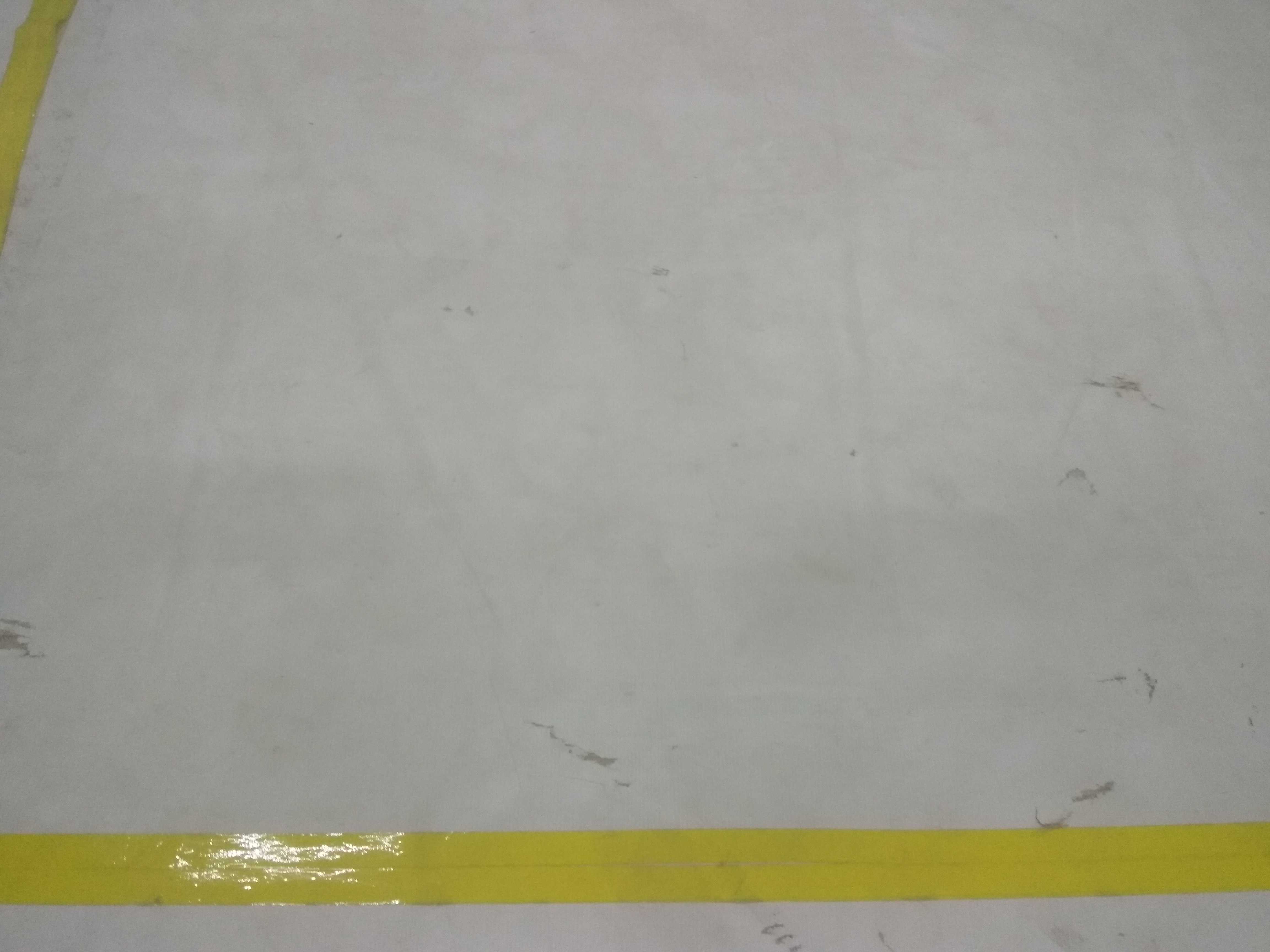}}}
  \caption{Offsets in the position of line with increasing roll angles}
\end{figure}

\section{STRATEGY}
While it is important to understand the environment and localize oneself, it is of equal importance to formulate a strategy for accomplishing the set goal. Our navigation stack helps the drone to comply with this simple notion. Like any other robotic decision-making process, we defined the following key modules on which our system works.

	\quad a. Perception 
    
    \quad b. Control and Localization
    
    \quad c. Strategy
    
Sections IV and V appertain to the first two modules. Depending on the environment, the strategy employed by the drone for its functioning can be modified. For example, if the grid structure is available, grid nodes are used as intermediate targets. In our implementation, these intermediate targets were identified and tracked by the strategy module depending upon the global target or task to be accomplished, like detection and localization of the goods using the bar-codes and QR codes. 
\subsection{Turn detection}
	Apart from nodes and vertical lines, curved or sharp turns can also be present in the environment which can be mistaken as a node if not dealt with properly by the algorithm. To avoid such a situation, an L-detection algorithm was implemented which differentiated an L from a regular node. The ratio of the number of pixels on the left to right side of the skeletonized line was considered, and a threshold ratio was set which, when exceeded, referred to a turn in the path.

\subsection{Multiple Line Handling}
The case of multiple nodes being detected in a given position is handled by tracking the goal node. Once the lines and their $\rho$ and $\theta$ values are detected, lines are divided into two different topics, vertical and horizontal. Vertical line published is used for the traversal of the drone. An imaginary horizontal line is assumed in front of the horizontal lines which helps the drone steer forwards. Multiple horizontal lines may cause problems when the height is changing as $\rho$ and $\theta$ value of each line might change significantly thus drone's returning to the same line becomes a difficult task. To tackle the problem, we have a constantly updating line coordinate which keeps track of current drone hover line parameters and updates the specific lines at different heights so that the drone doesn't lose the tracking in case multiple horizontal lines are visible. The strategy commands the controller to hover the drone over a specific goal node. We also included some exceptional cases that are bound to occur, e.g., the starting point/take-off strategy and endpoint/charge station detection for actual deployment scenario. 


\begin{figure*}[t]
\begin{center}
a. \includegraphics[width=1in,height=1in]{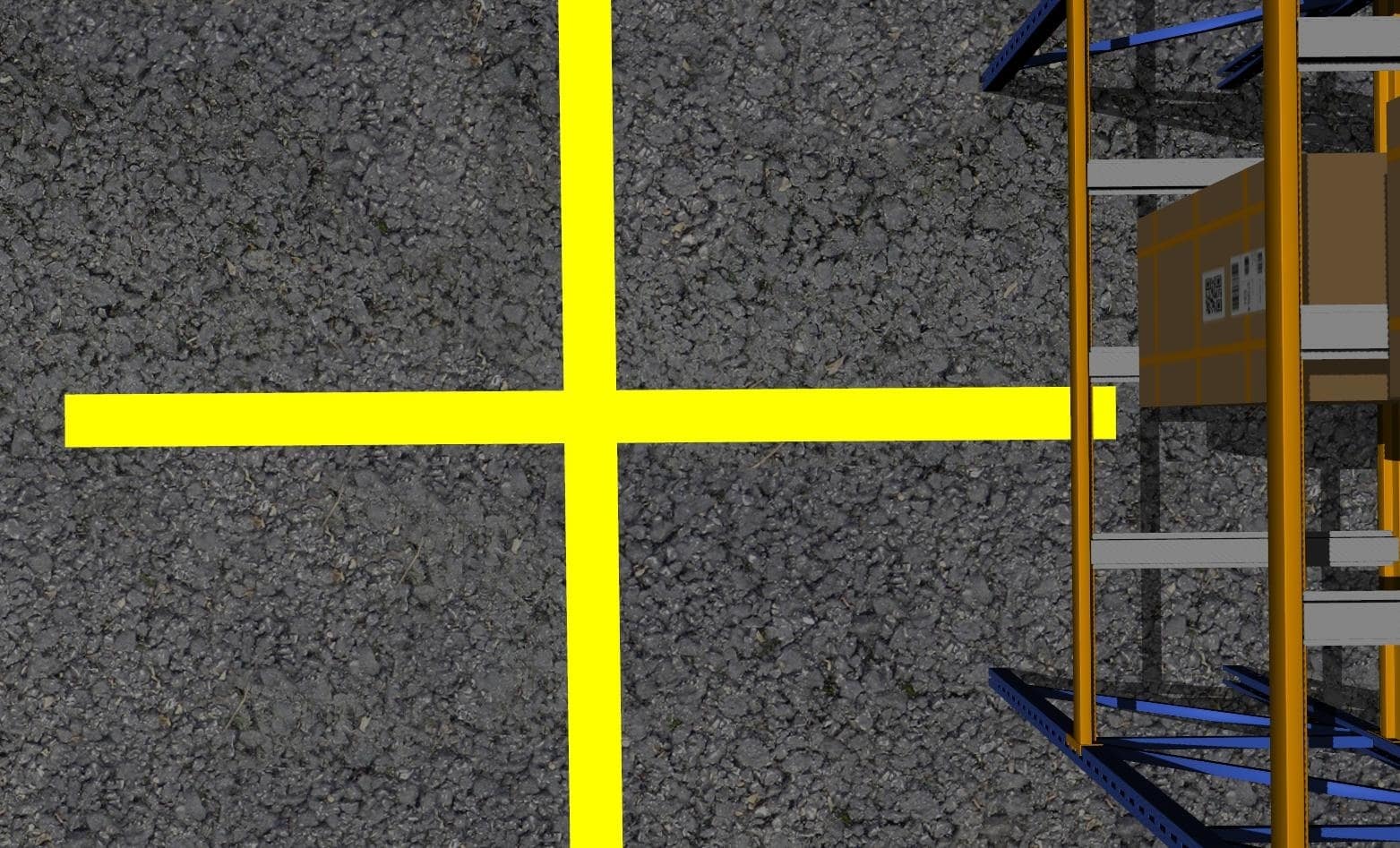}
\hspace{0.1cm}
\includegraphics[width=1in,height=1in]{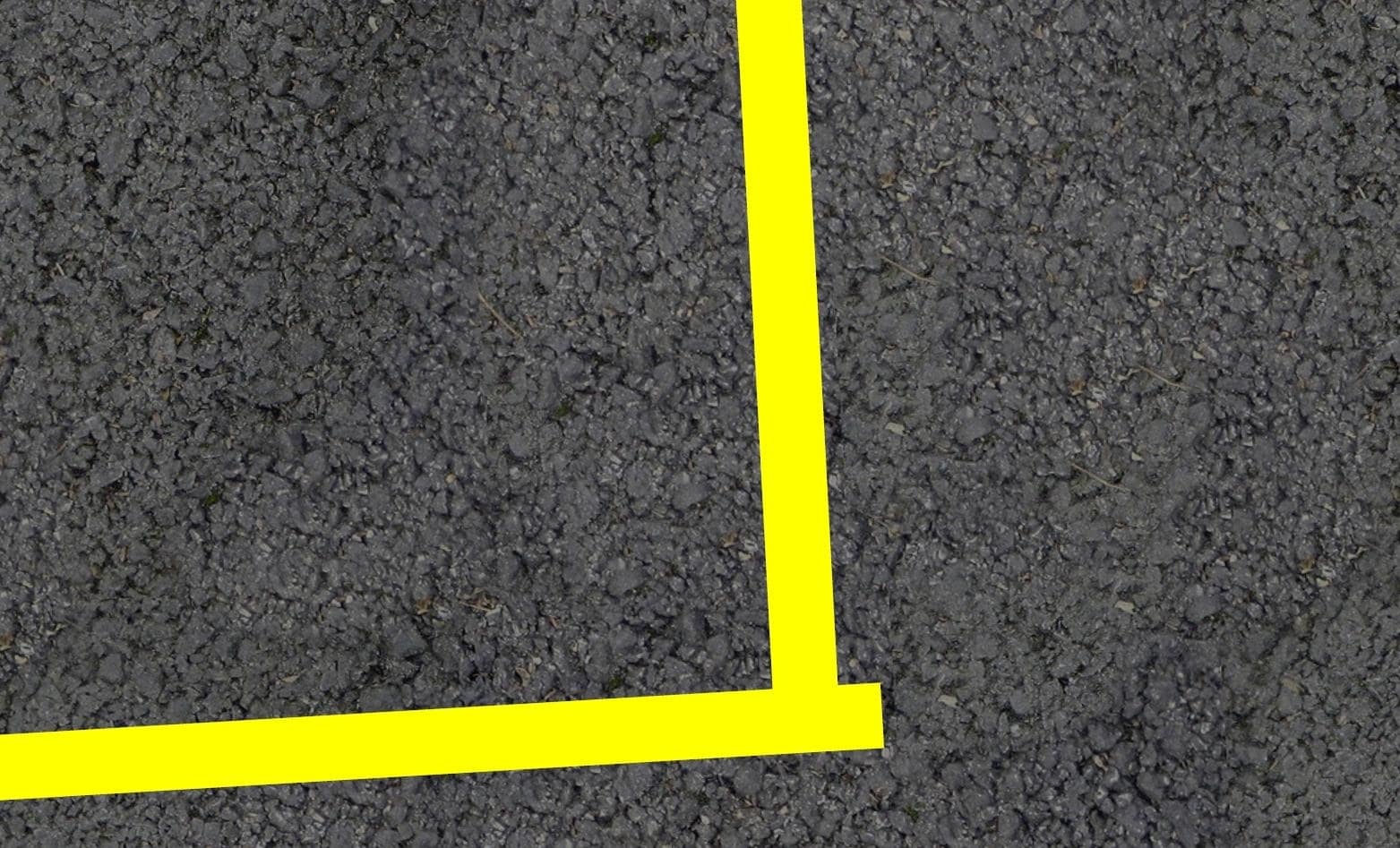}
\hspace{0.1cm}
\includegraphics[width=1in,height=1in]{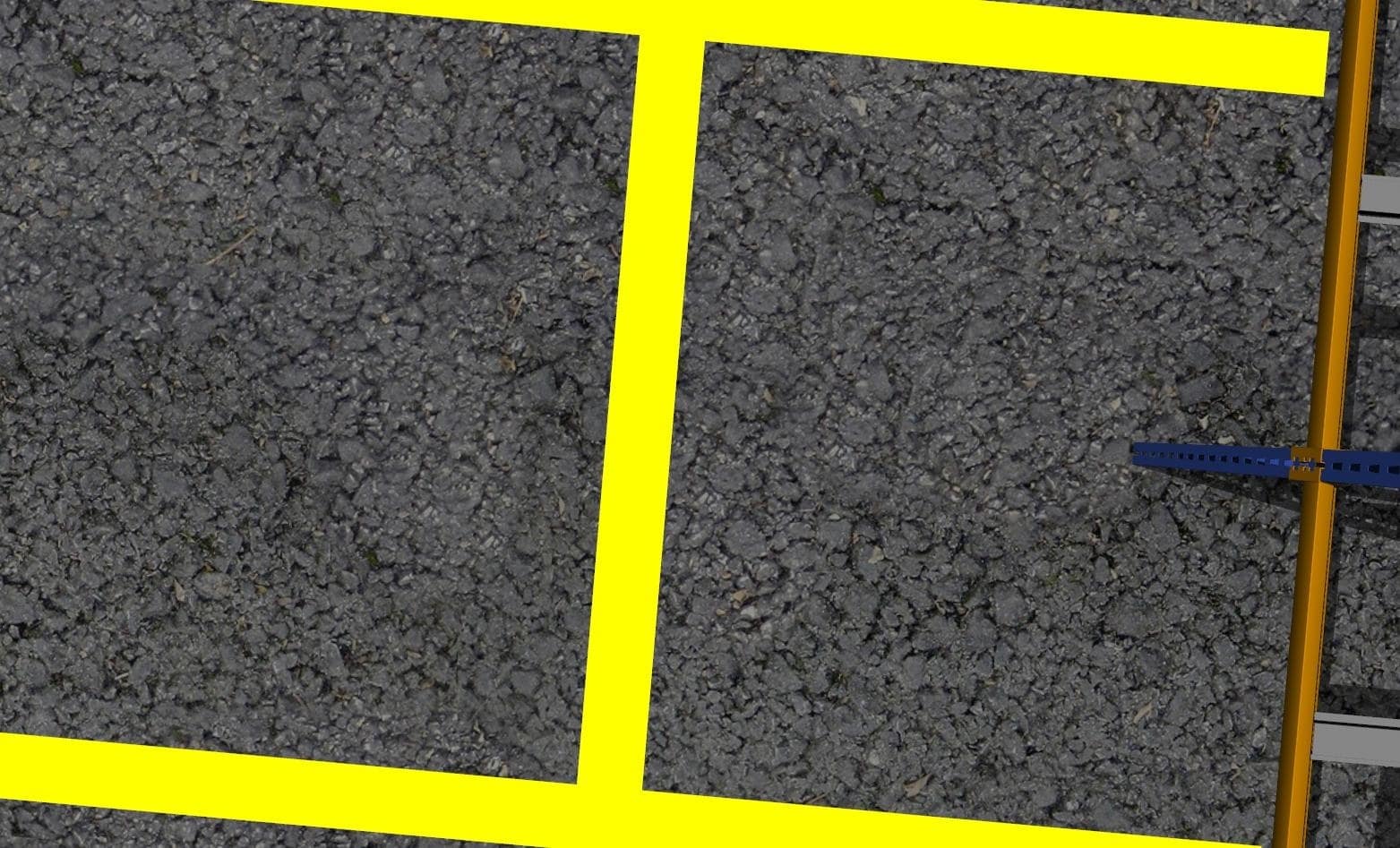}
\hspace{0.1cm}
\includegraphics[width=1in,height=1in]{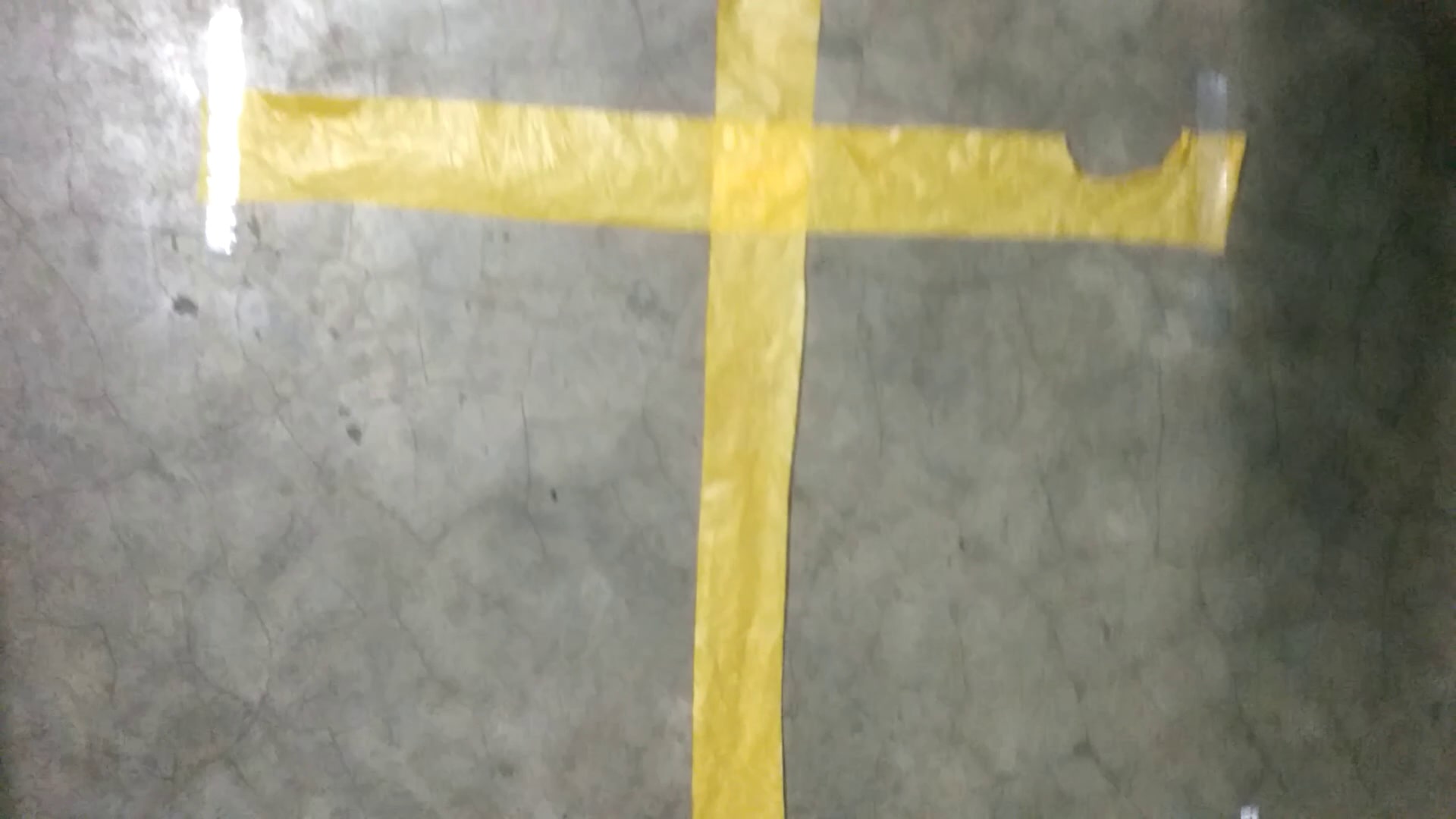}
\hspace{0.1cm}
\includegraphics[width=1in,height=1in]{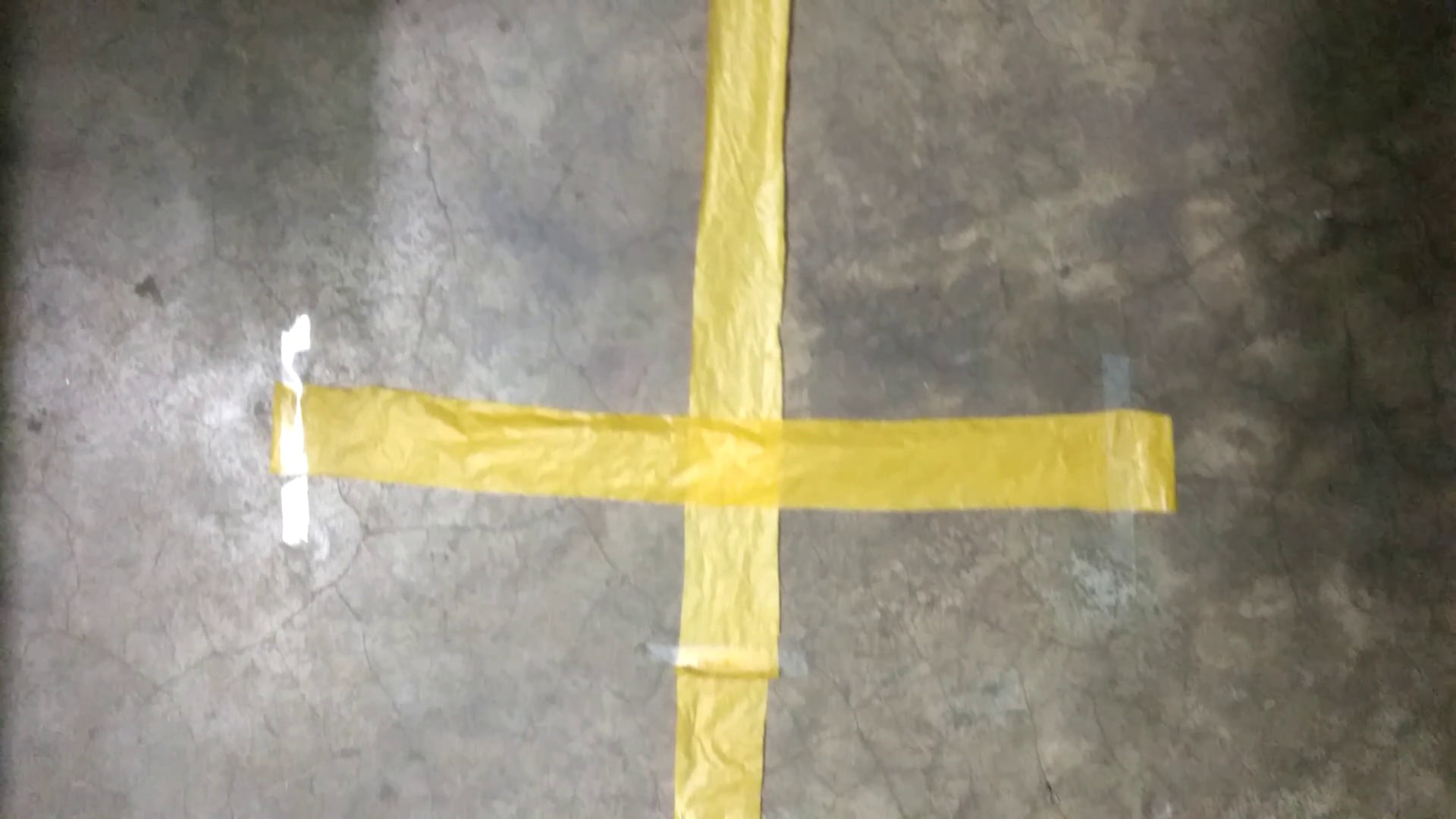}
\hspace{0.1cm}
\includegraphics[width=1in,height=1in]{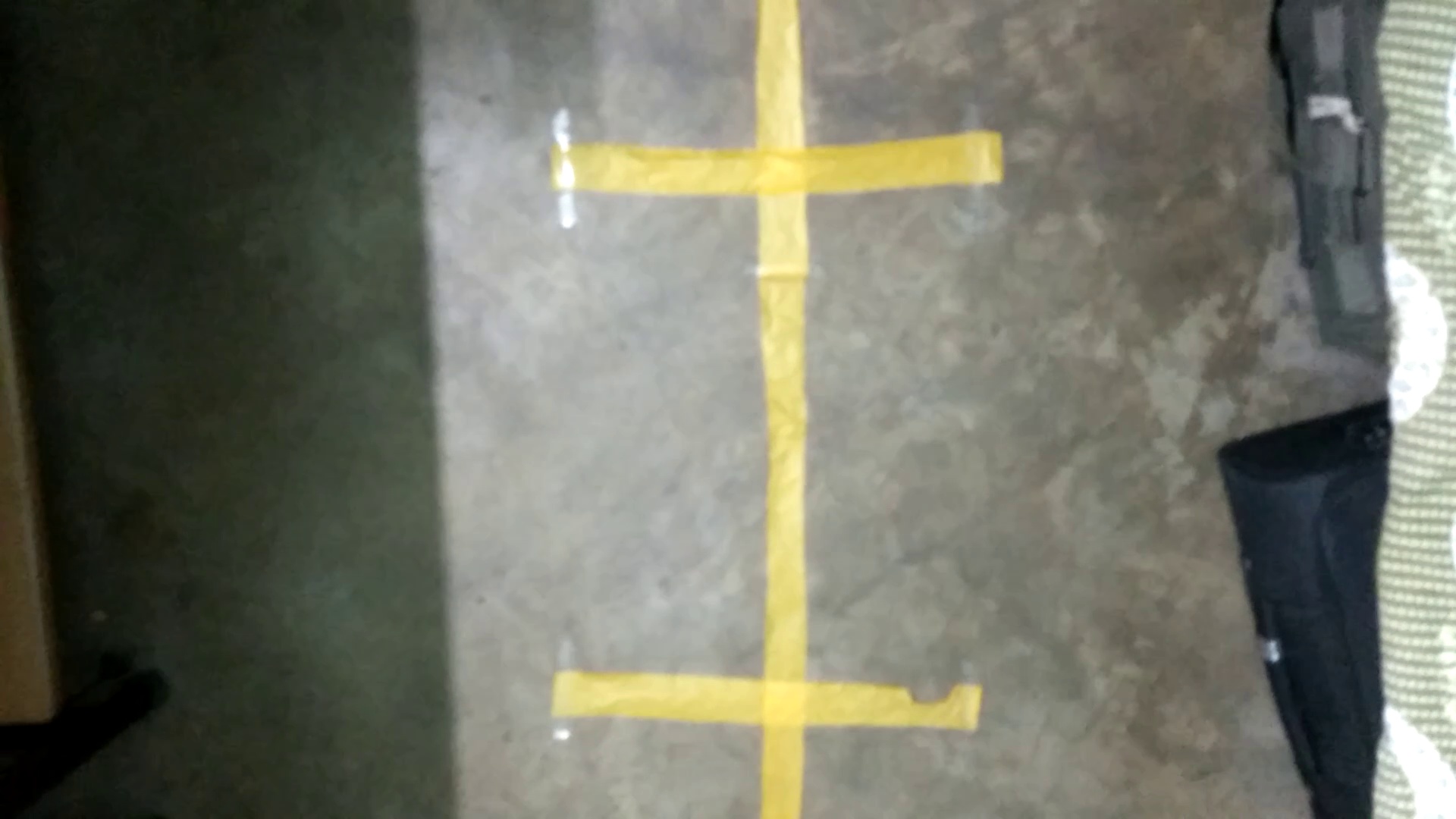}
\end{center}

\begin{center}
b. \includegraphics[width=1in,height=1in]{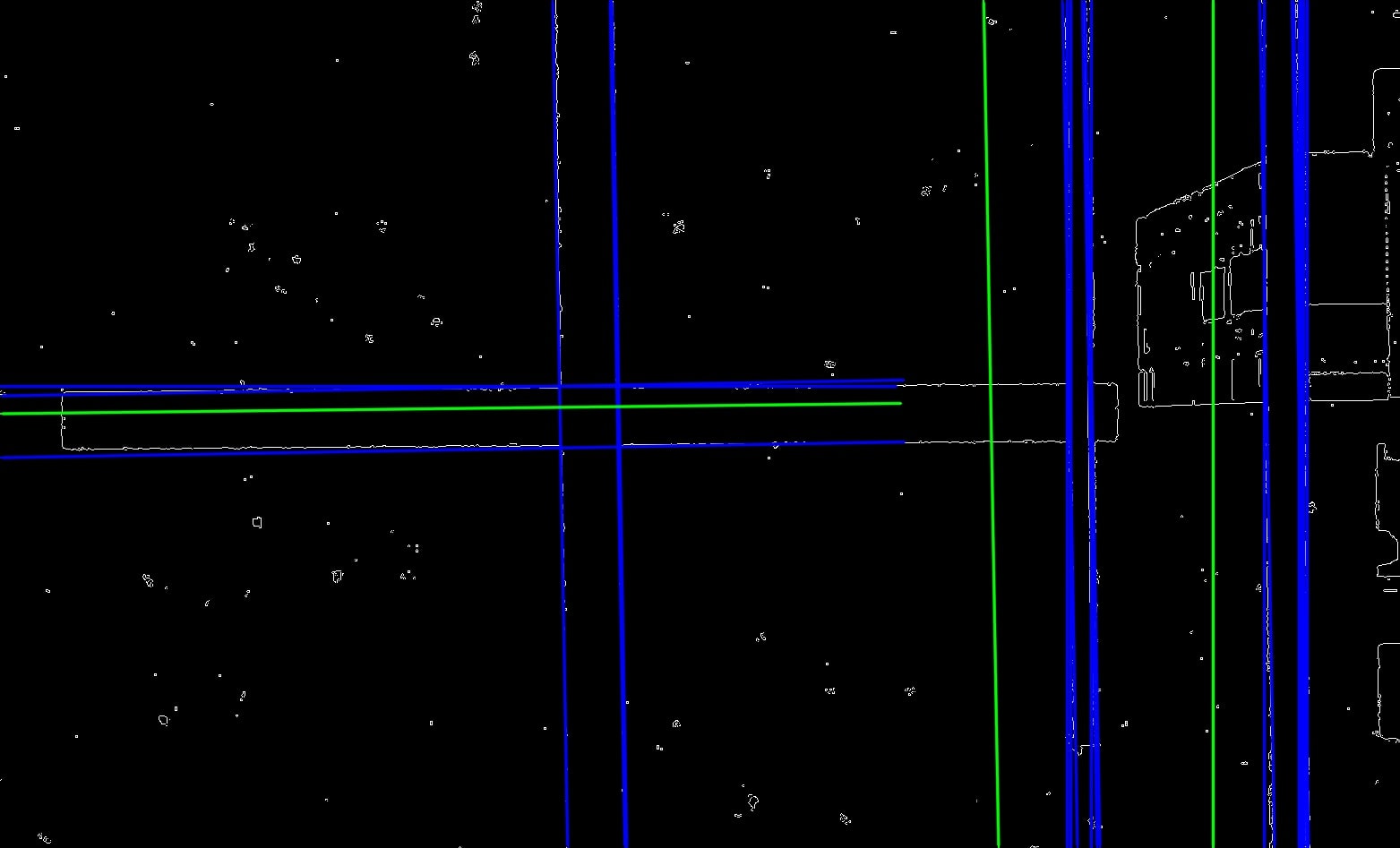}
\hspace{0.1cm}
\includegraphics[width=1in,height=1in]{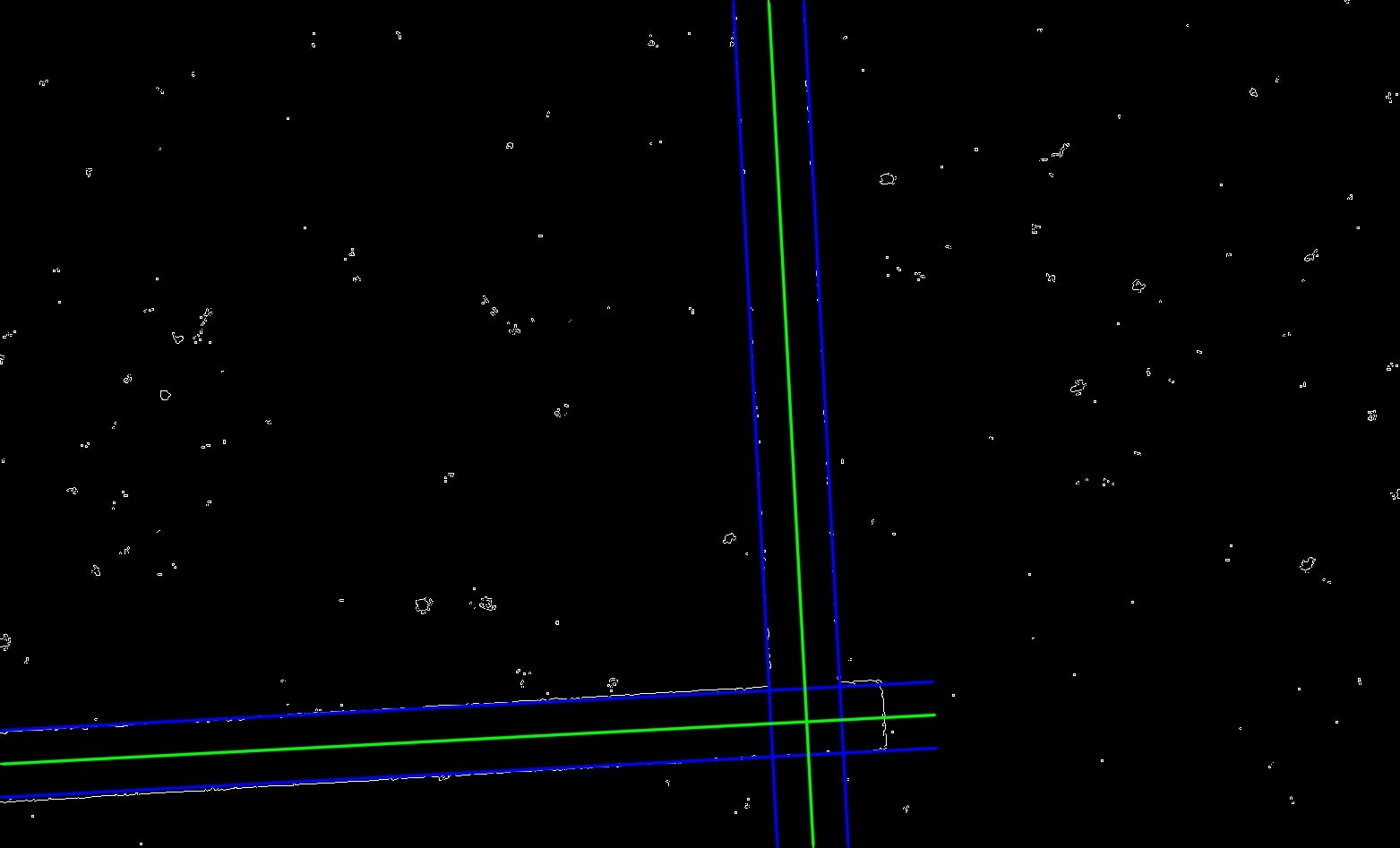}
\hspace{0.1cm}
\includegraphics[width=1in,height=1in]{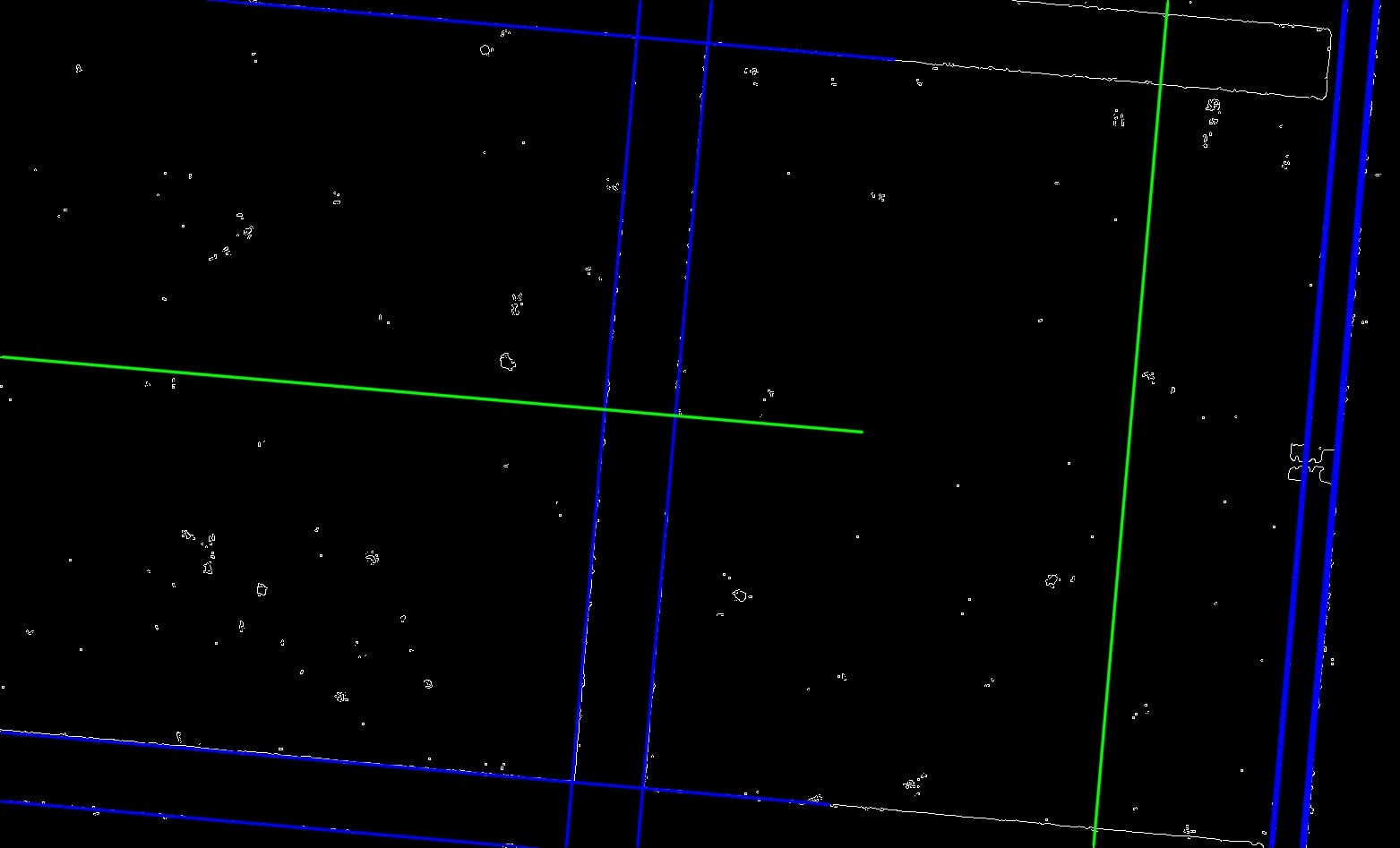}
\hspace{0.1cm}
\includegraphics[width=1in,height=1in]{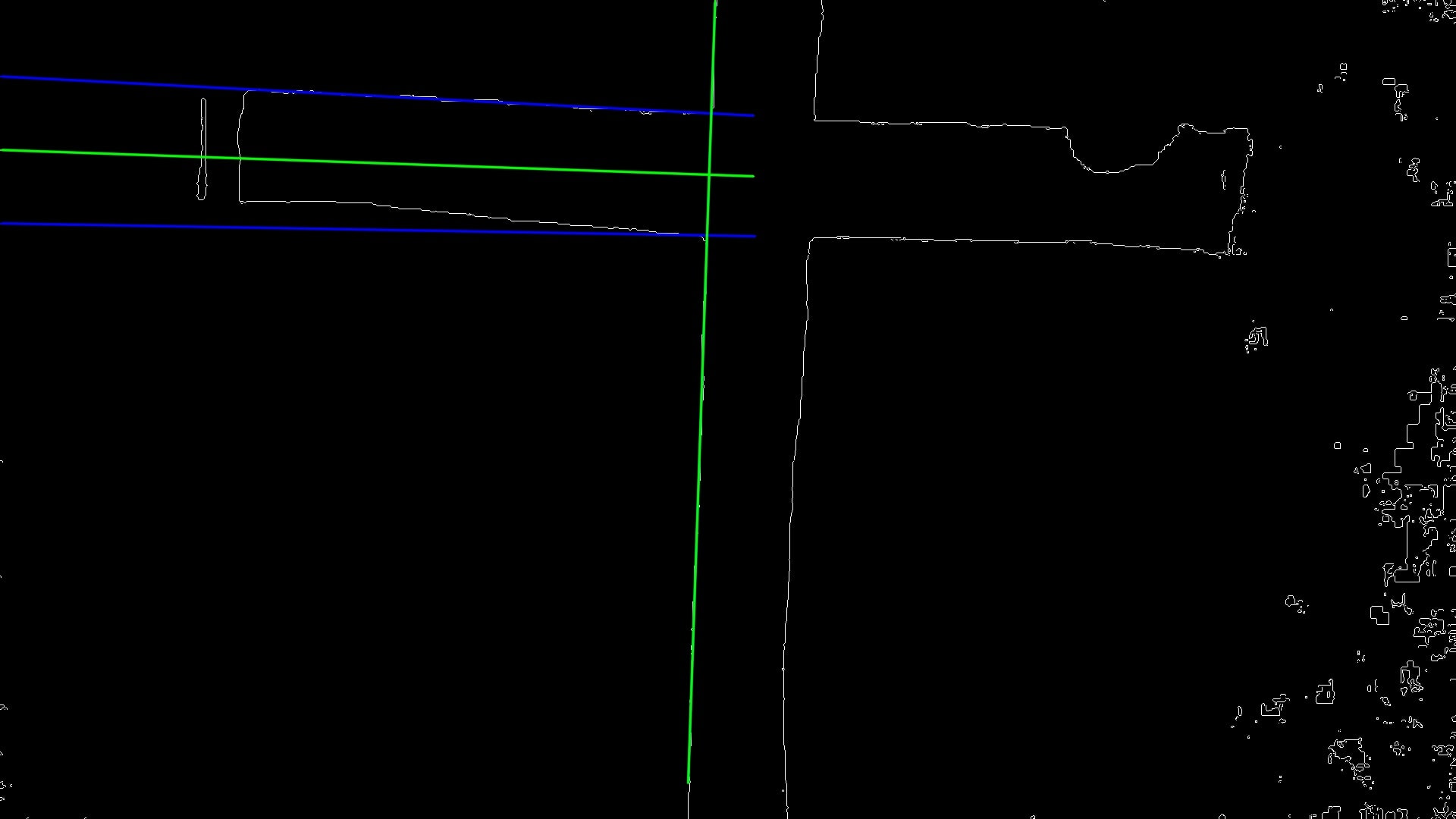}
\hspace{0.1cm}
\includegraphics[width=1in,height=1in]{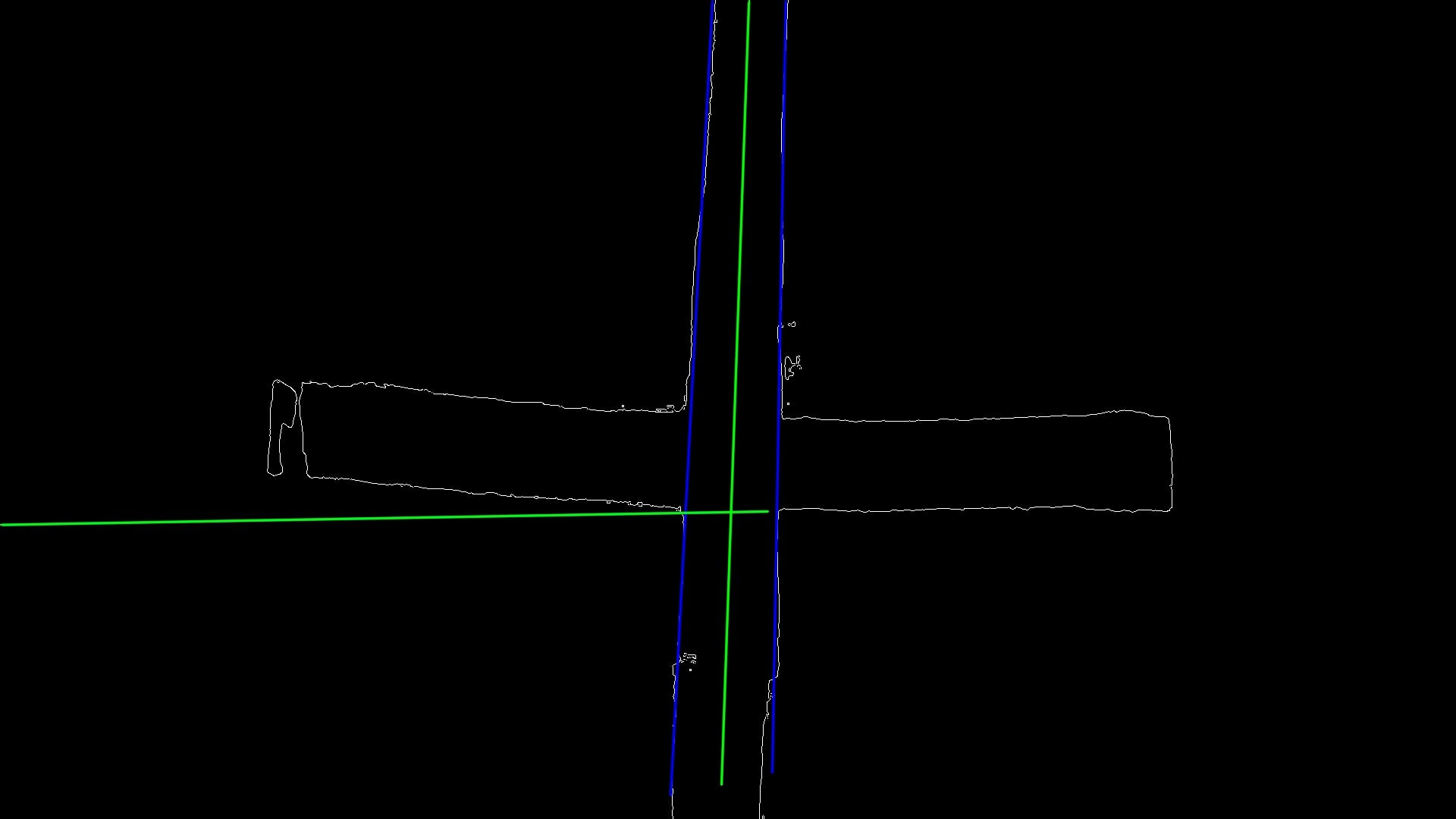}
\hspace{0.1cm}
\includegraphics[width=1in,height=1in]{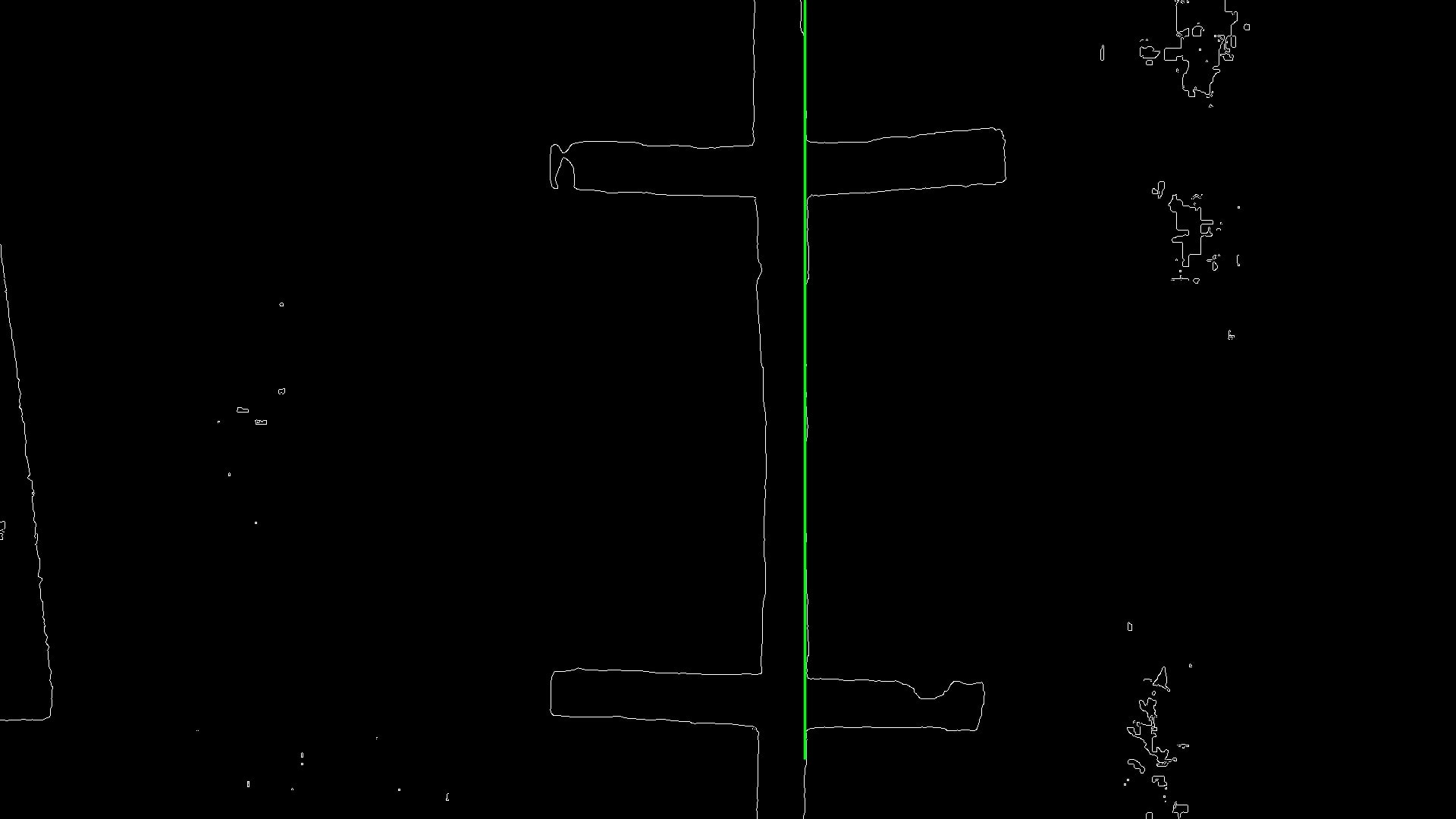}
\end{center}

\begin{center}
c. \includegraphics[width=1in,height=1in]{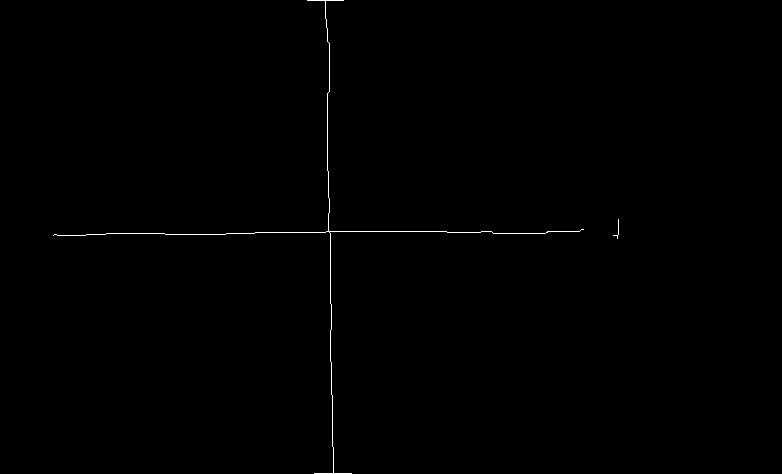}
\hspace{0.1cm}
\includegraphics[width=1in,height=1in]{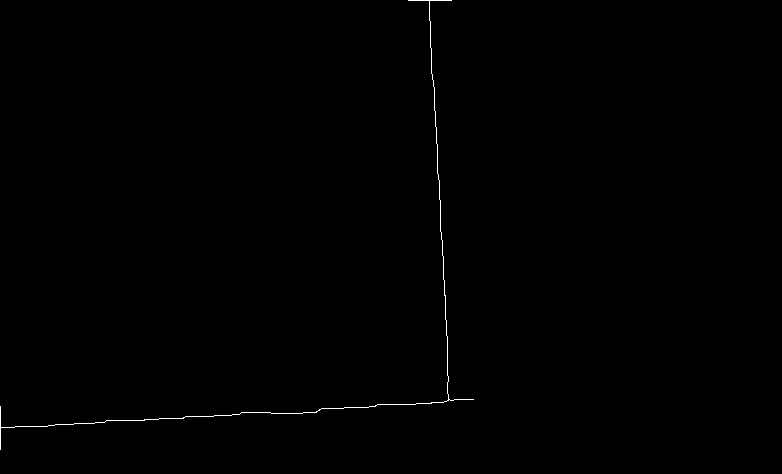}
\hspace{0.1cm}
\includegraphics[width=1in,height=1in]{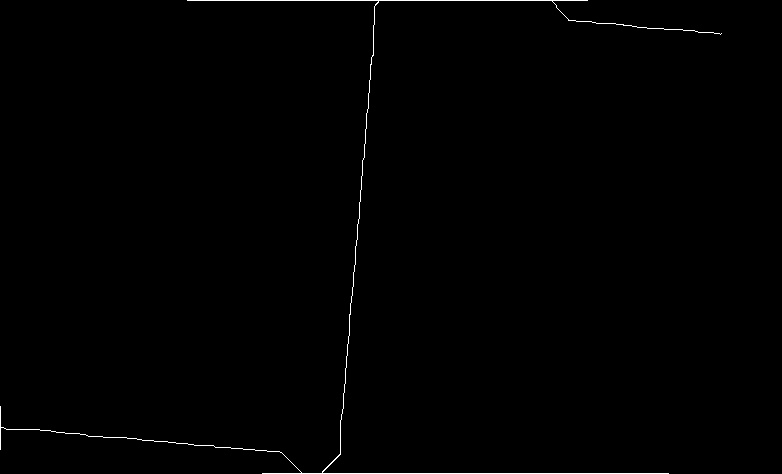}
\hspace{0.1cm}
\includegraphics[width=1in,height=1in]{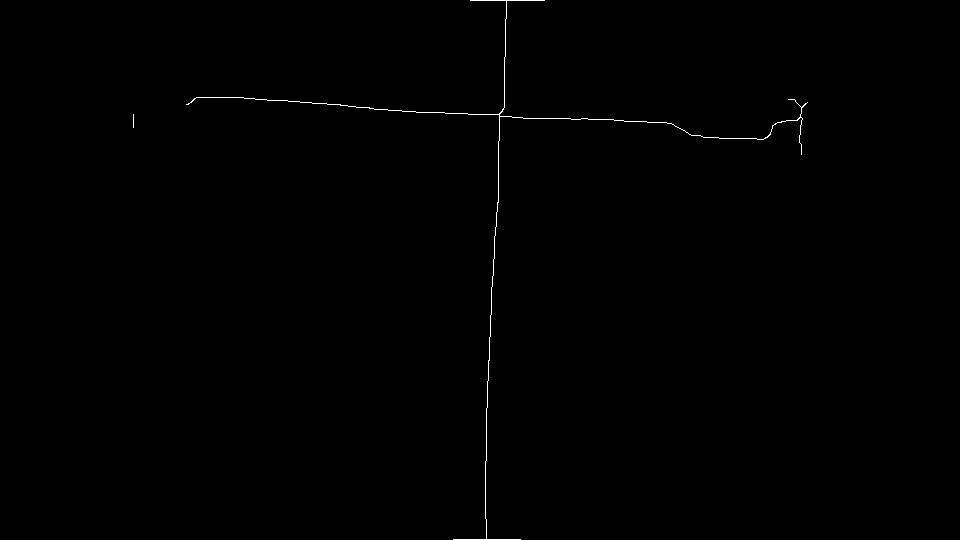}
\hspace{0.1cm}
\includegraphics[width=1in,height=1in]{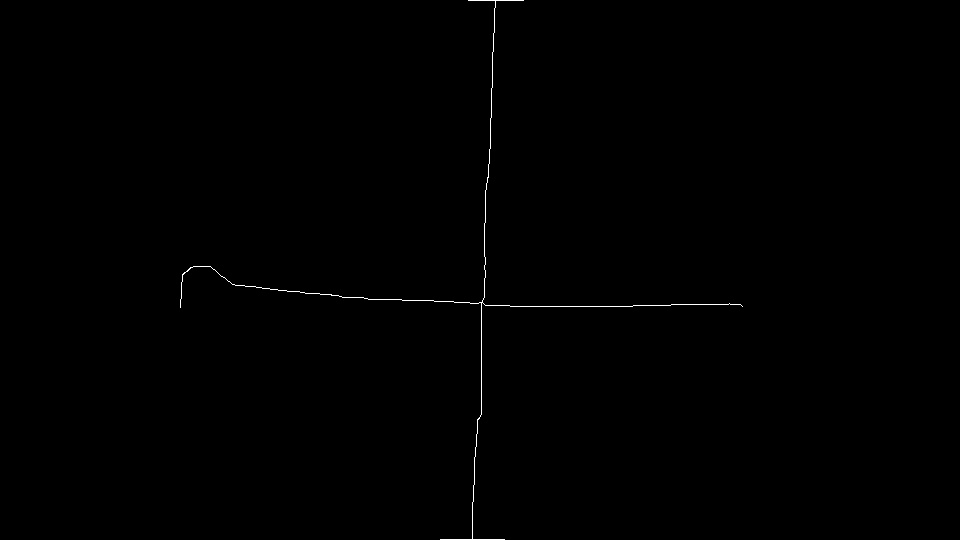}
\hspace{0.1cm}
\includegraphics[width=1in,height=1in]{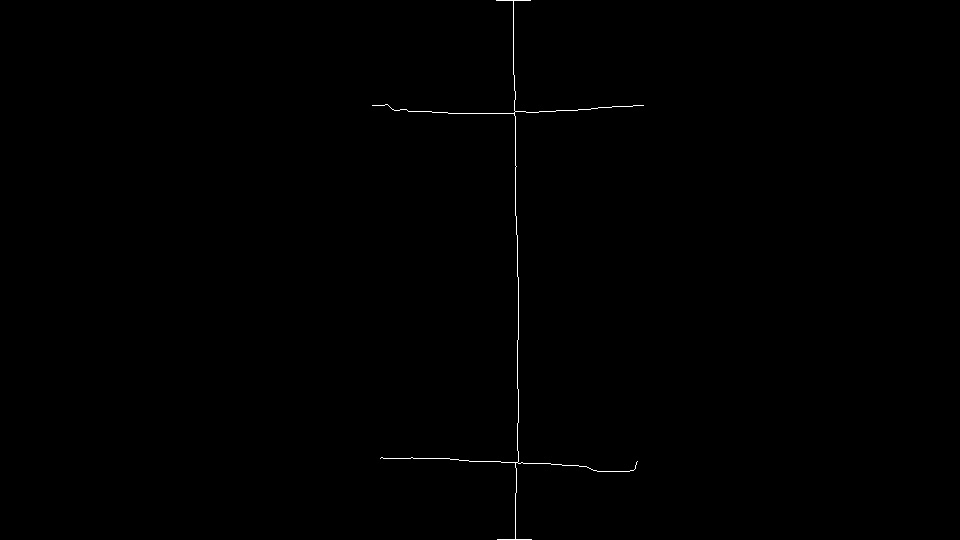}
\end{center}

\begin{center}
d. \includegraphics[width=1in,height=1in]{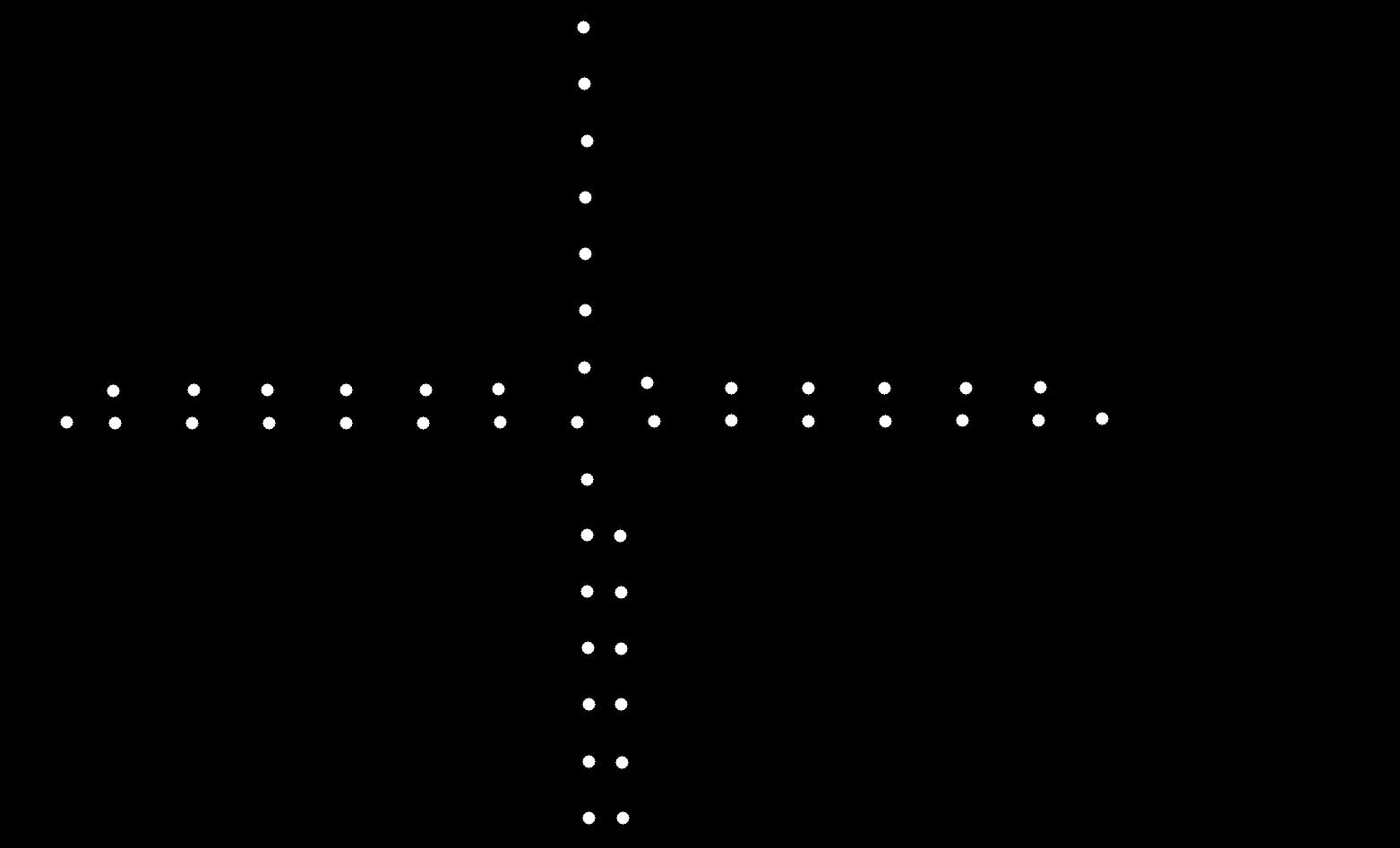}
\hspace{0.1cm}
\includegraphics[width=1in,height=1in]{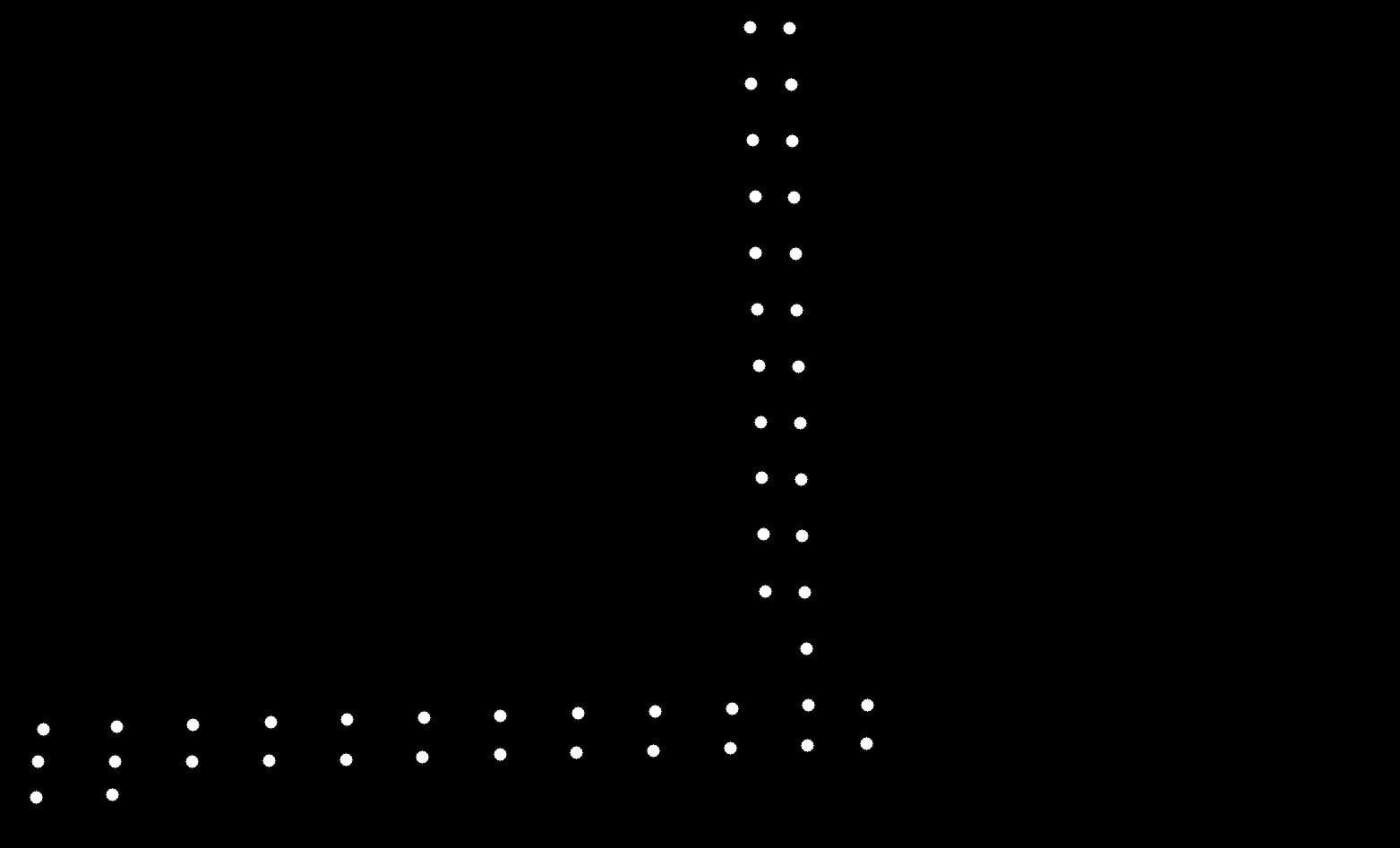}
\hspace{0.1cm}
\includegraphics[width=1in,height=1in]{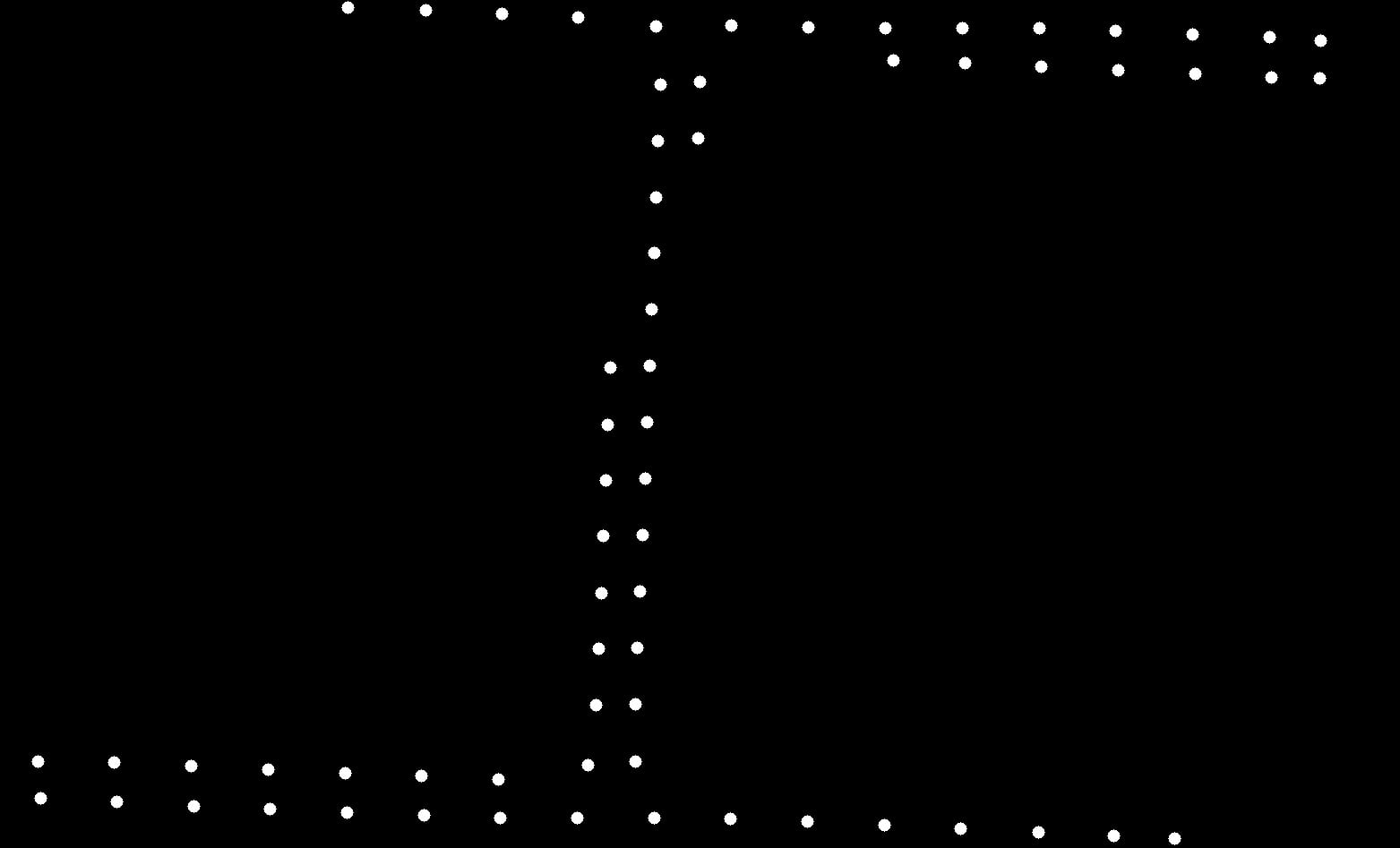}
\hspace{0.1cm}
\includegraphics[width=1in,height=1in]{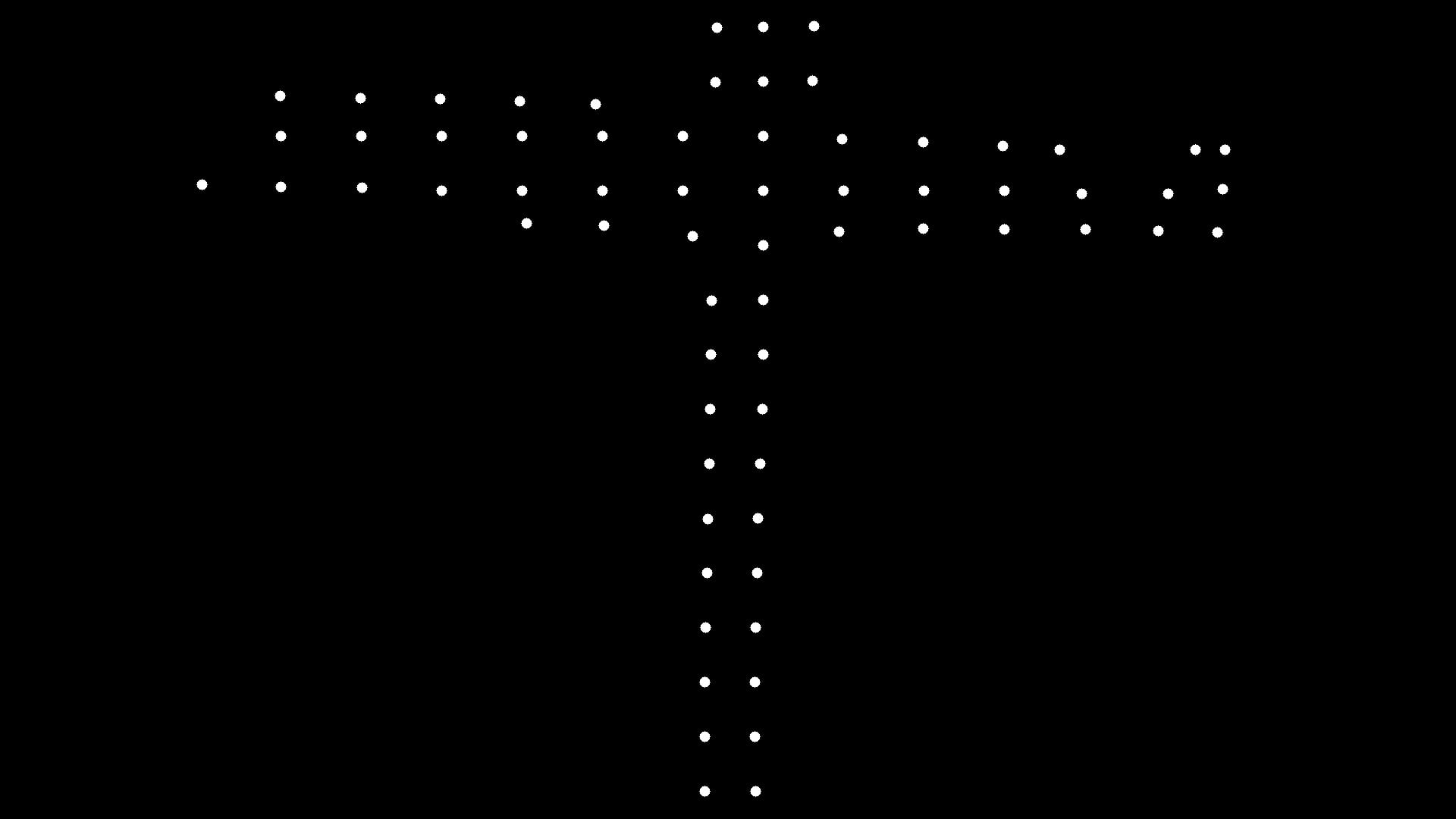}
\hspace{0.1cm}
\includegraphics[width=1in,height=1in]{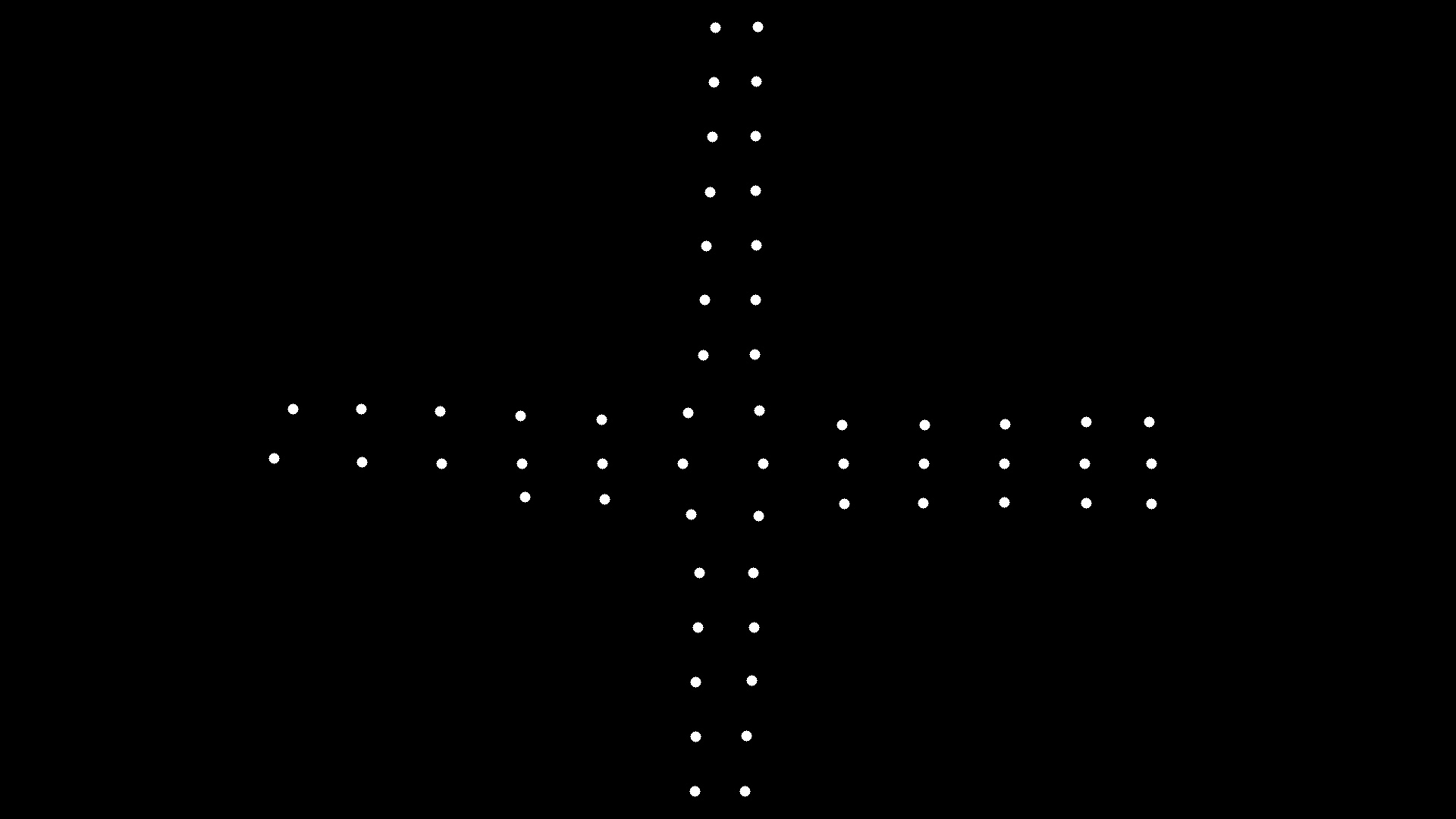}
\hspace{0.1cm}
\includegraphics[width=1in,height=1in]{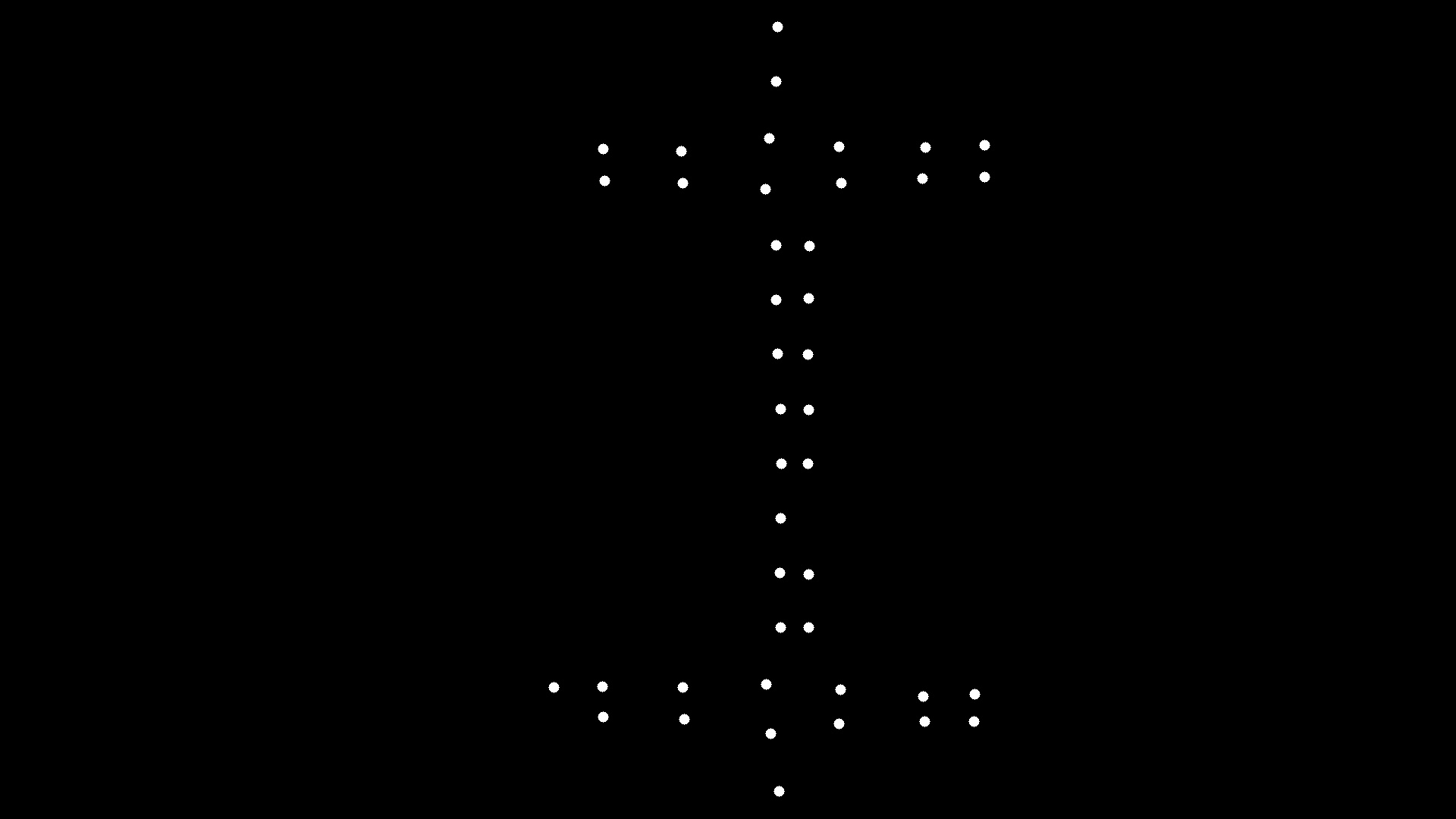}
\end{center}

\begin{center}
e. \includegraphics[width=1in,height=1in]{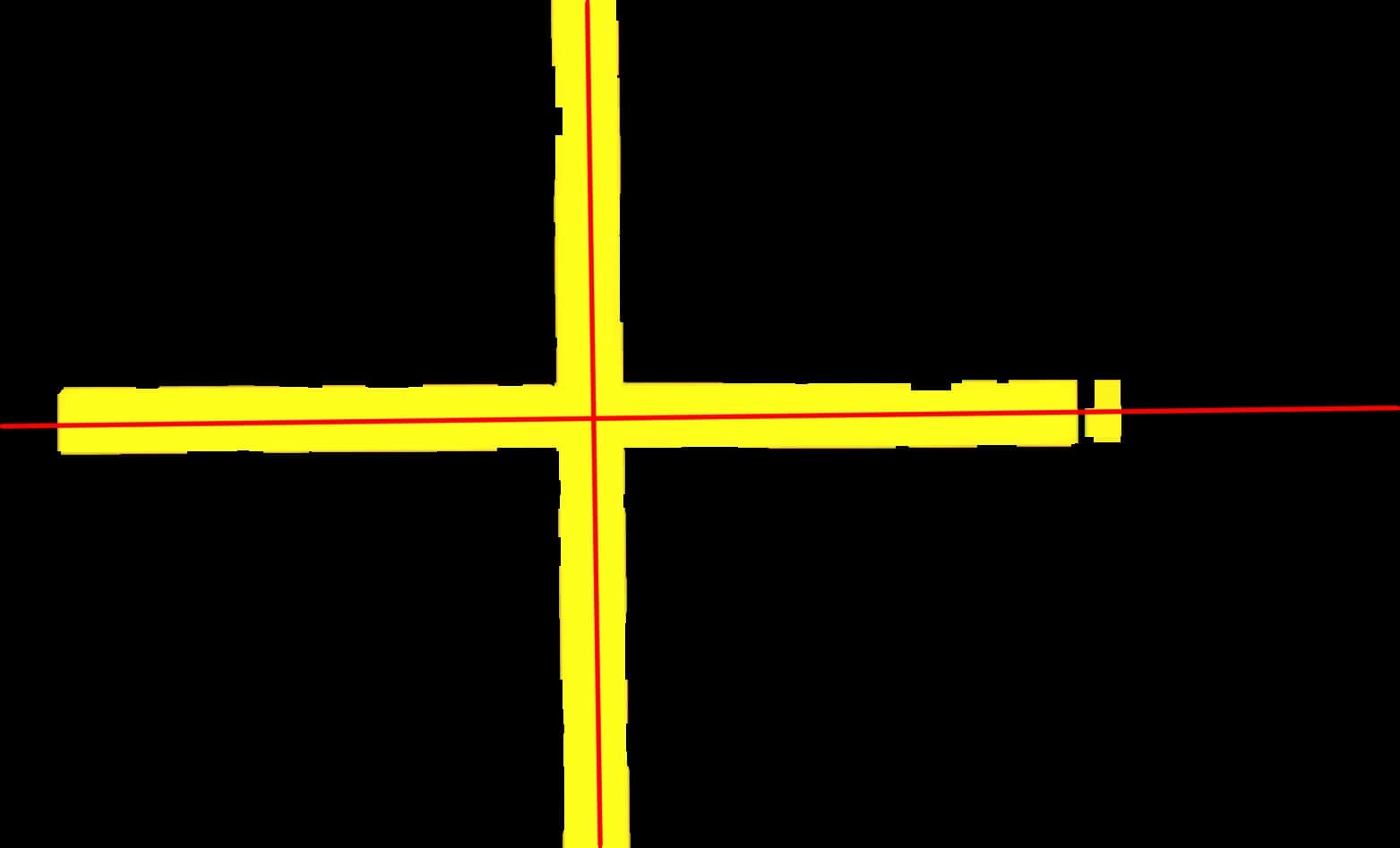}
\hspace{0.1cm}
\includegraphics[width=1in,height=1in]{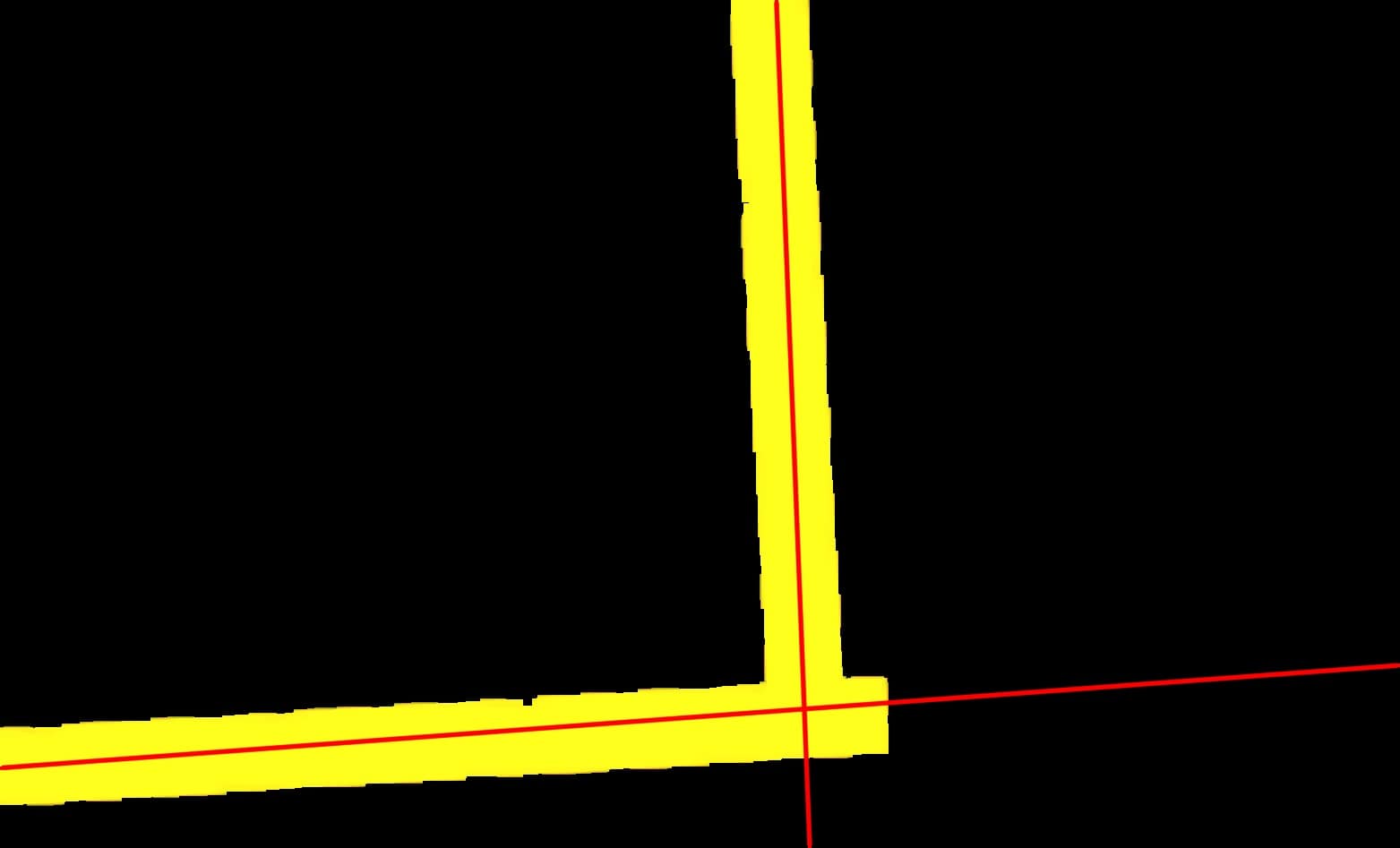}
\hspace{0.1cm}
\includegraphics[width=1in,height=1in]{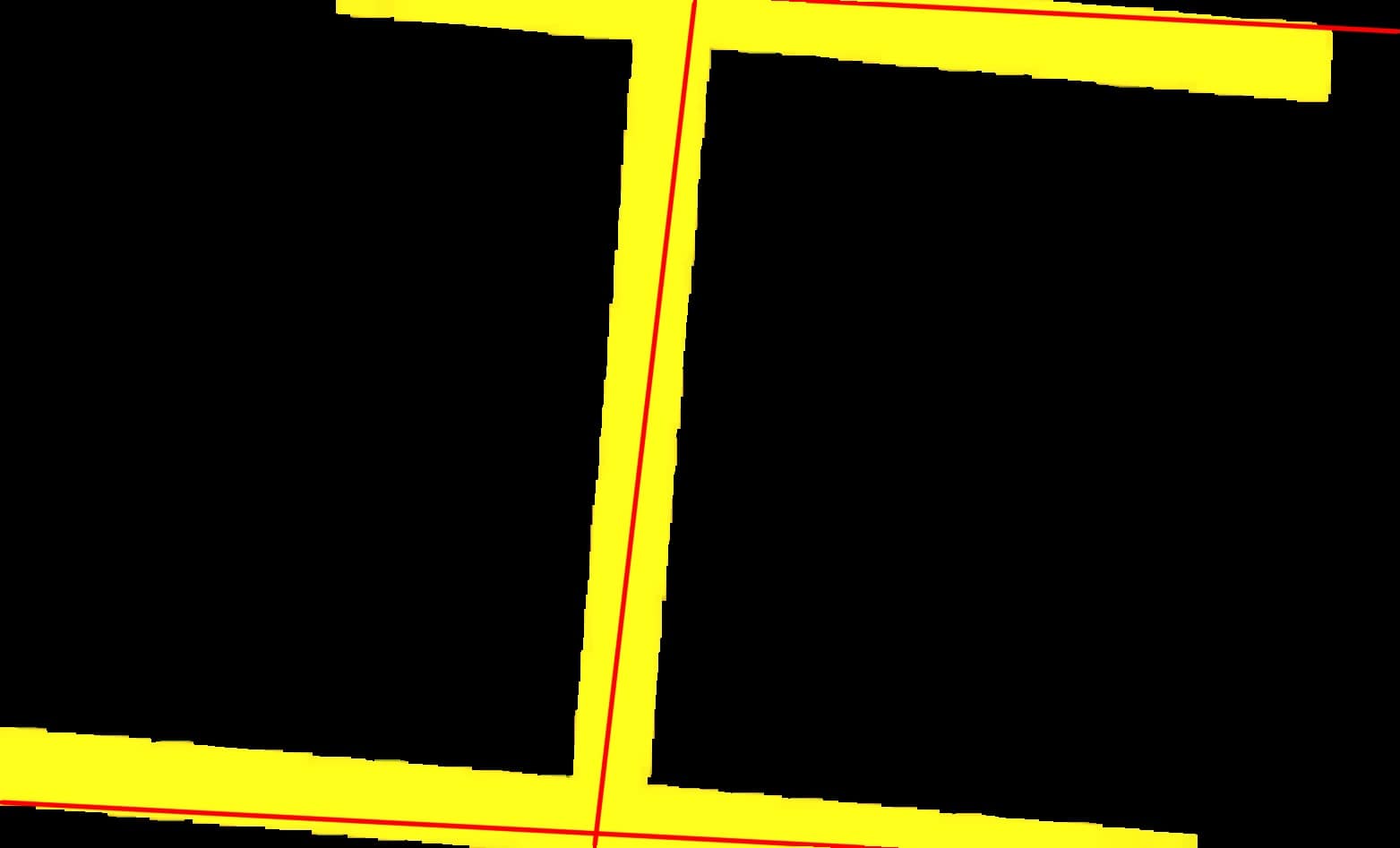}
\hspace{0.1cm}
\includegraphics[width=1in,height=1in]{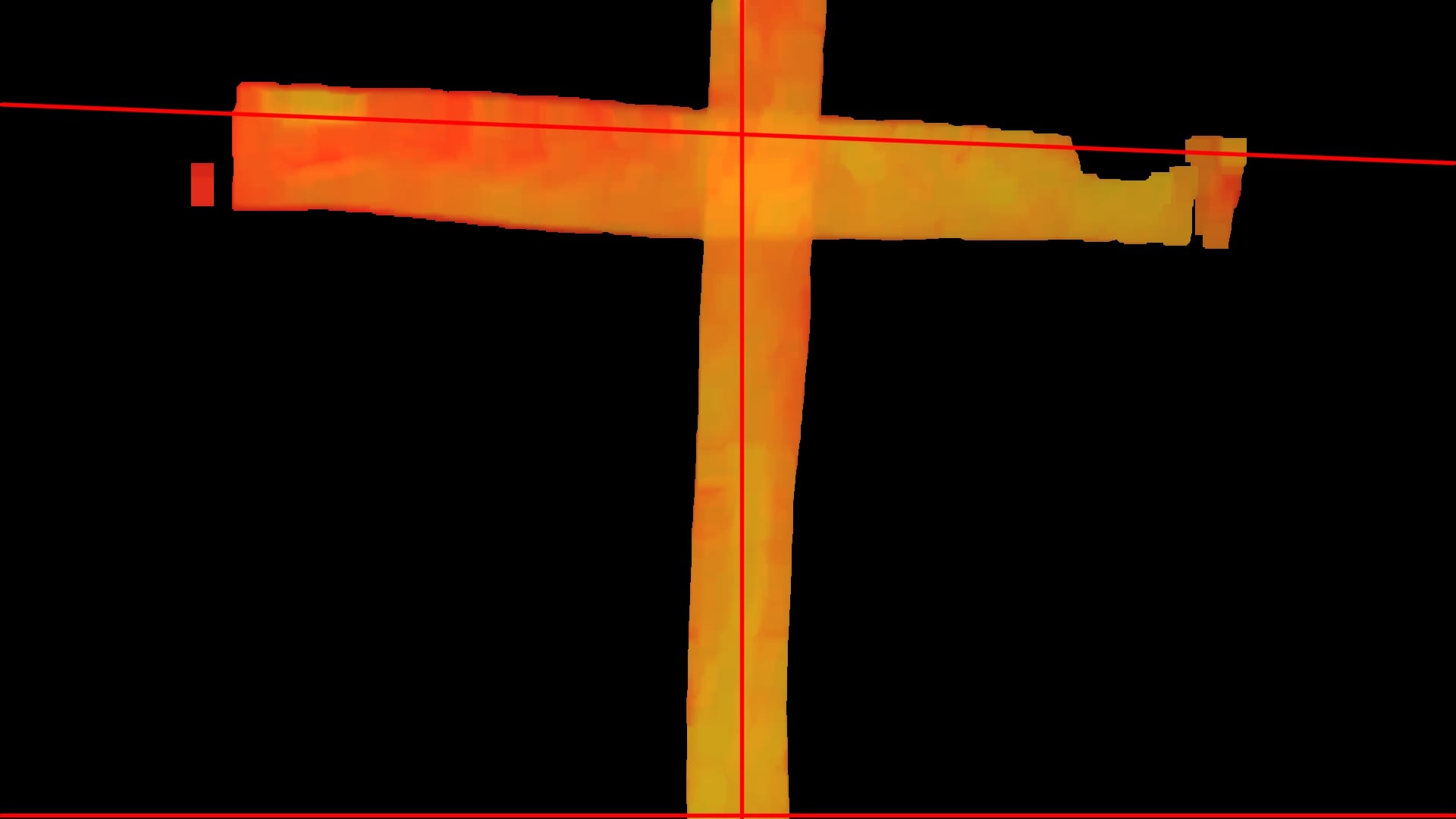}
\hspace{0.1cm}
\includegraphics[width=1in,height=1in]{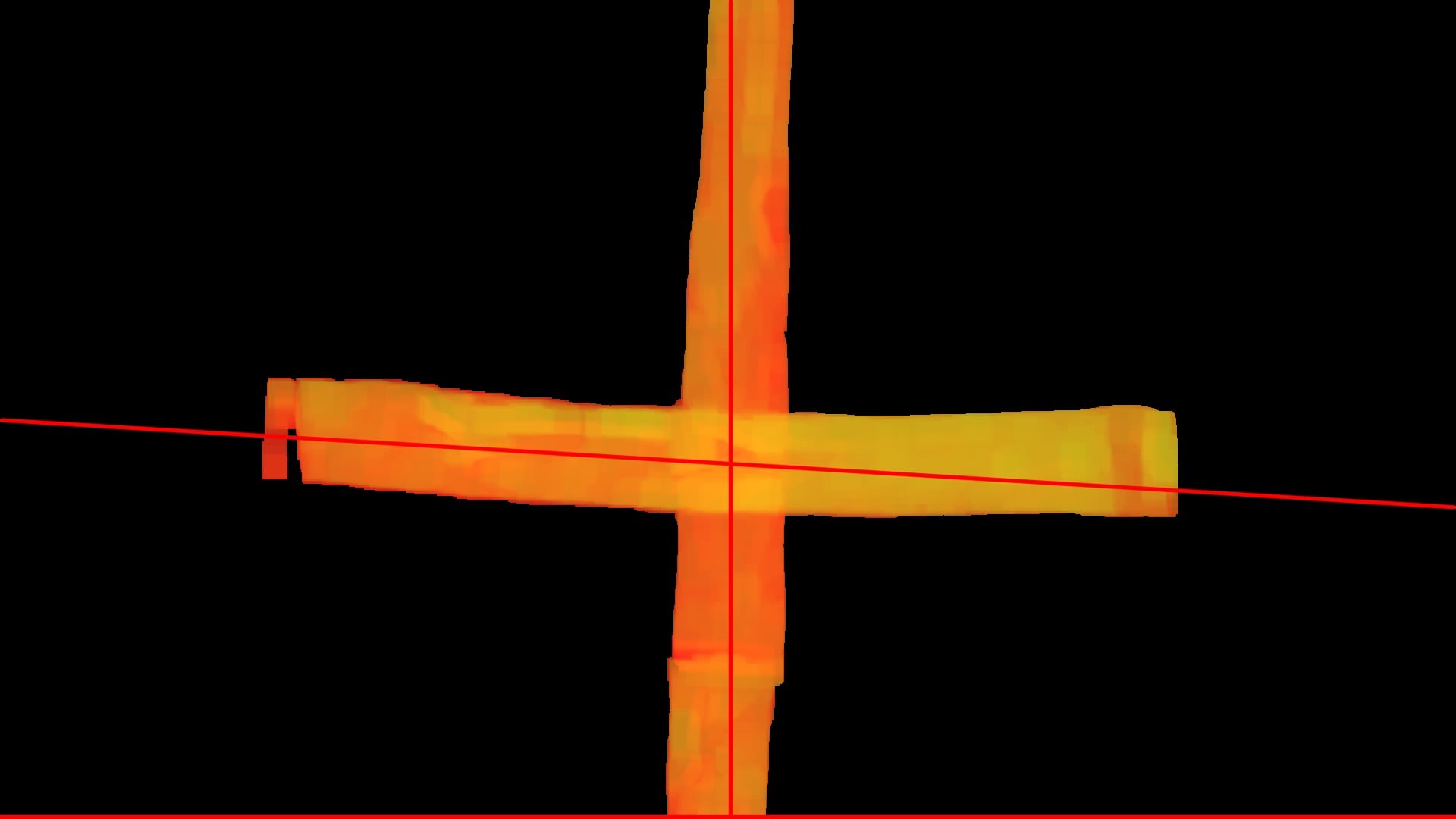}
\hspace{0.1cm}
\includegraphics[width=1in,height=1in]{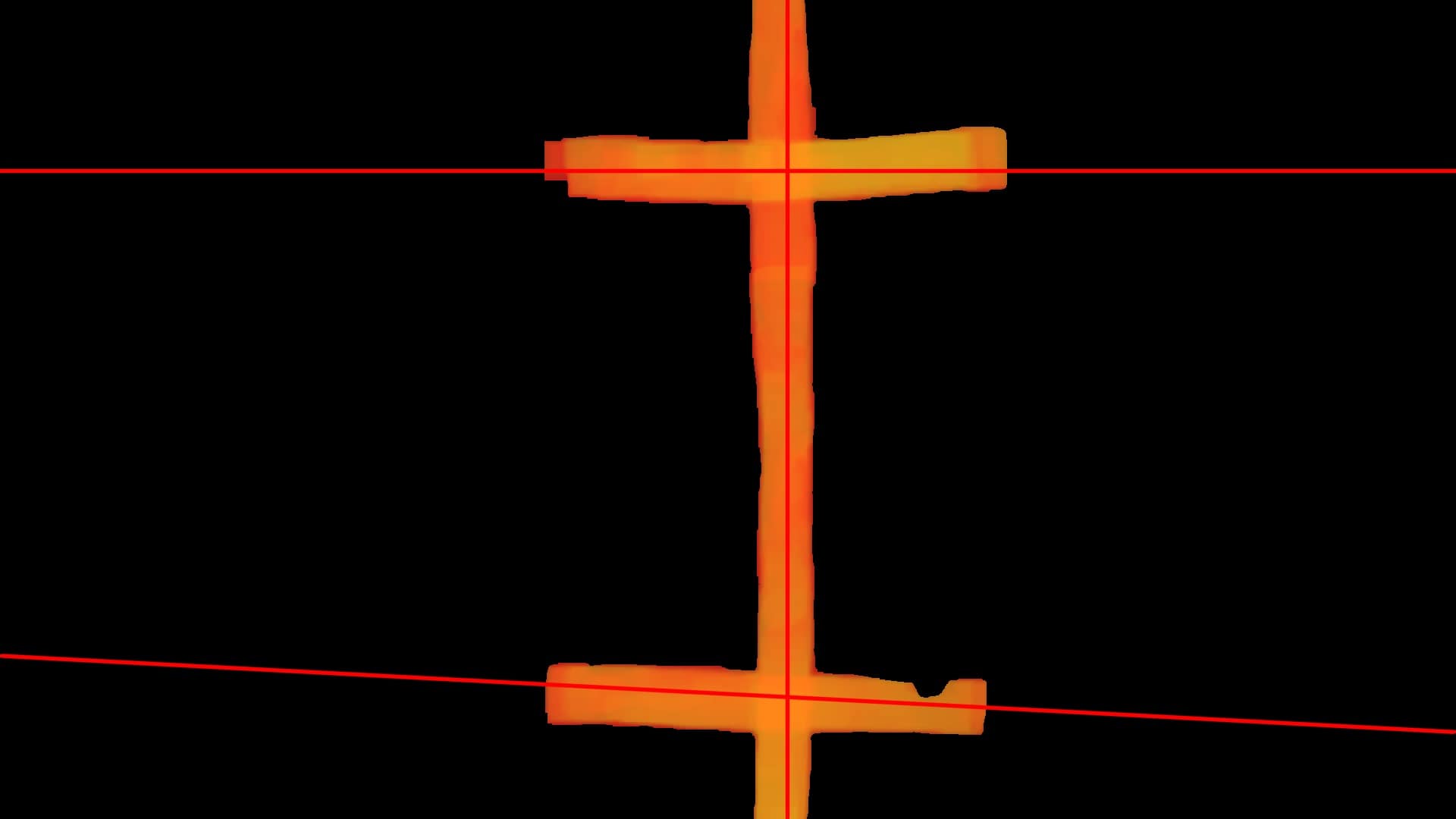}
\end{center}

\begin{center}
    \caption{From Top to Bottom: a. Test Cases of the warehouse strip;  b. Line detection using naive hough line, canny and manual thresholding; c. Skeleton generation using the thinning algorithm; d. Output from centroid method; e. Merged final output from skeletonization and centroid approach.  }
\end{center}
\end{figure*}


\section{EXPERIMENTAL SETUP AND EVALUATION}
All the demonstrated methods and strategies were implemented on a Parrot AR Drone 2 with variants like off-board processing on remote compute unit and on-board processing with Raspberry Pi 3. The remote compute unit was equipped with Intel Core i7 7th Gen @2.7GHz x 4 processor with 32 GB on-board RAM. For the other variant, Raspberry Pi 3 was attached to the bottom of AR Drone which led to a decrease in flight time as it drew power from the drone's battery and also added up to the drone's weight. This was compensated using a Lithium-ion battery that provided more charge density than the standard one. We implemented our pipeline on these variants and evaluated the performance. Before the actual hardware deployment, all the necessary tests were made on a Gazebo simulation as shown in Figure 13. AR drone hosts two cameras, from which we targeted the down-facing camera for localization and the front-facing camera for detecting QR-Code/Barcode. We chose Robot Operating System (ROS) for retrieving drone's on-board sensors data and also for sending appropriate command actions to it. ROS package ardrone\_autonomy was used for this purpose. Image processing on drone-sent camera data was done with the help of the OpenCV library. Finally, all the data handling and processing codes were written in optimized multi-thread C++ code to match the industrial standards. Table I shows On-board (fps) Vs Off-board (fps) for Centroid and Skeletonization Methods.

\subsection{Evaluation Metric and Results}
Better performance of the grid-localization relates to better grid-line detection by the machine vision algorithm. Hence, we provide an evaluation, particularly for the line detection algorithms. Any line in a plane can be represented by Hessian normal form of the equation of line in parameters $r$ (normal distance of the line from the origin) and $\alpha$ (angle made by the normal vector to the line with one of the axis). We measured the performance of different methods by calculating the absolute errors in $r$ and $\alpha$ of the generated line with respect to the ground truth line. We collected about 50 images containing different views of the grid structure both from the simulation world and real world experimental setup each. The ground truth for each of these images was drawn manually as perceived by the human mindset and was later converted into hessian normal form. Some of the images and the result of the application of discussed methods in the previous sections are shown in Figure 12. Table II shows the performance of methods on an average over 50 images. Results show Naive approach suffer due to thick lines. This causes it to create multiple lines and thus reducing the accuracy of the approach. Moreover, multiple horizontal lines are averaged out as one due to clustering which further causes error, especially in real test cases. Centroid approach works by dividing the line into smaller parts and taking average over it which reduces the errors significantly and provides very accurate $\rho$ and $\theta$ values. Skeletonization, on other hand, works by reducing the lines to single pixel thickness. This induces errors in $\theta$ of the final line but comparatively fewer errors in $\rho$ compared to centroid approach. Combined approach sees a balanced accuracy of both the approaches separately as it normalizes some of the drawbacks from each approach.

\begin{figure}[ht]
  \centering
  {{\includegraphics[width=3.8cm]{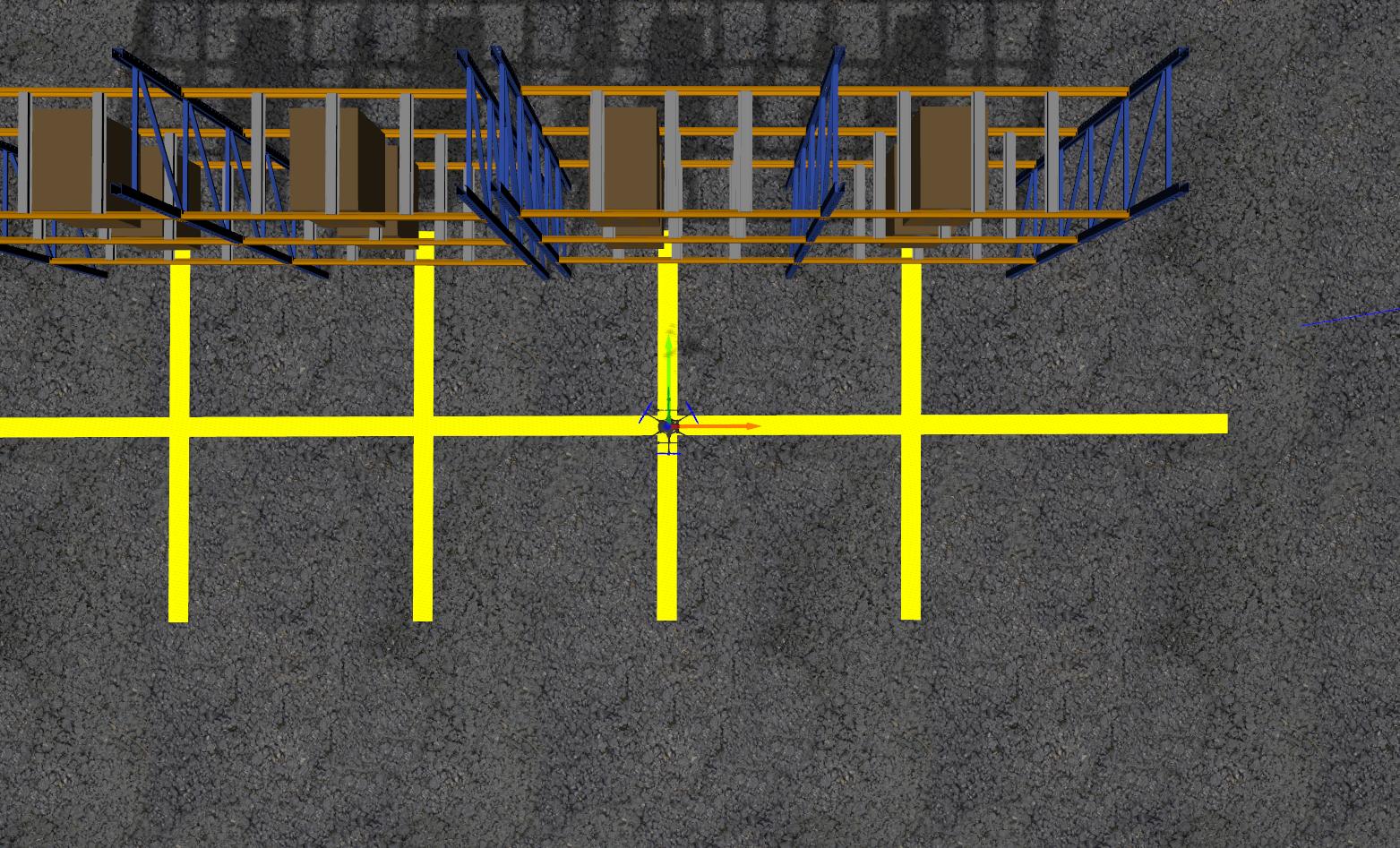}}}
  \qquad
  {{\includegraphics[width=3.8cm]{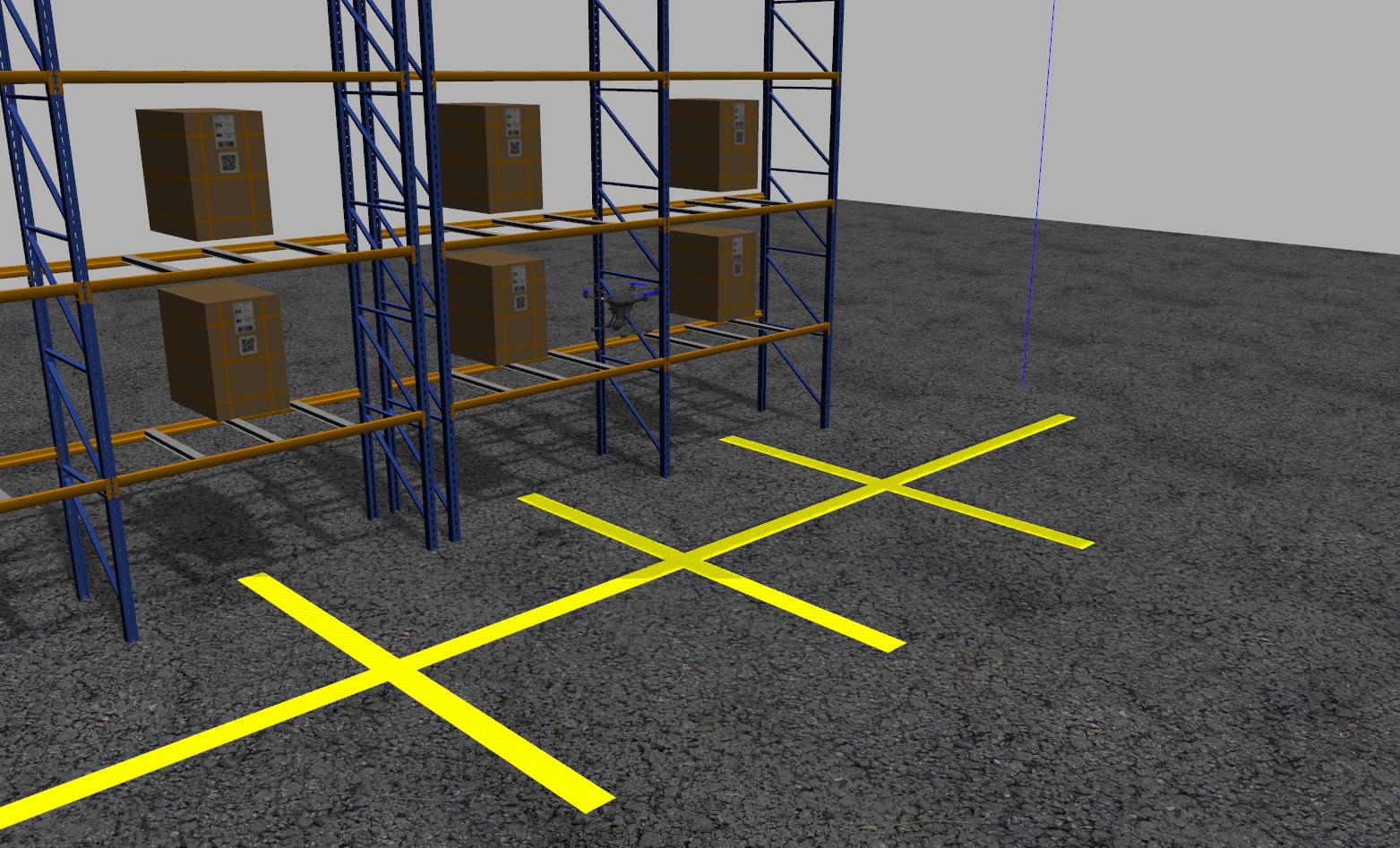}}}
  \caption{Simulation environment developed for verifying complete navigation stack}
\end{figure}

\begin{figure}[ht]
  \centering
  {{\includegraphics[width=5.0cm]{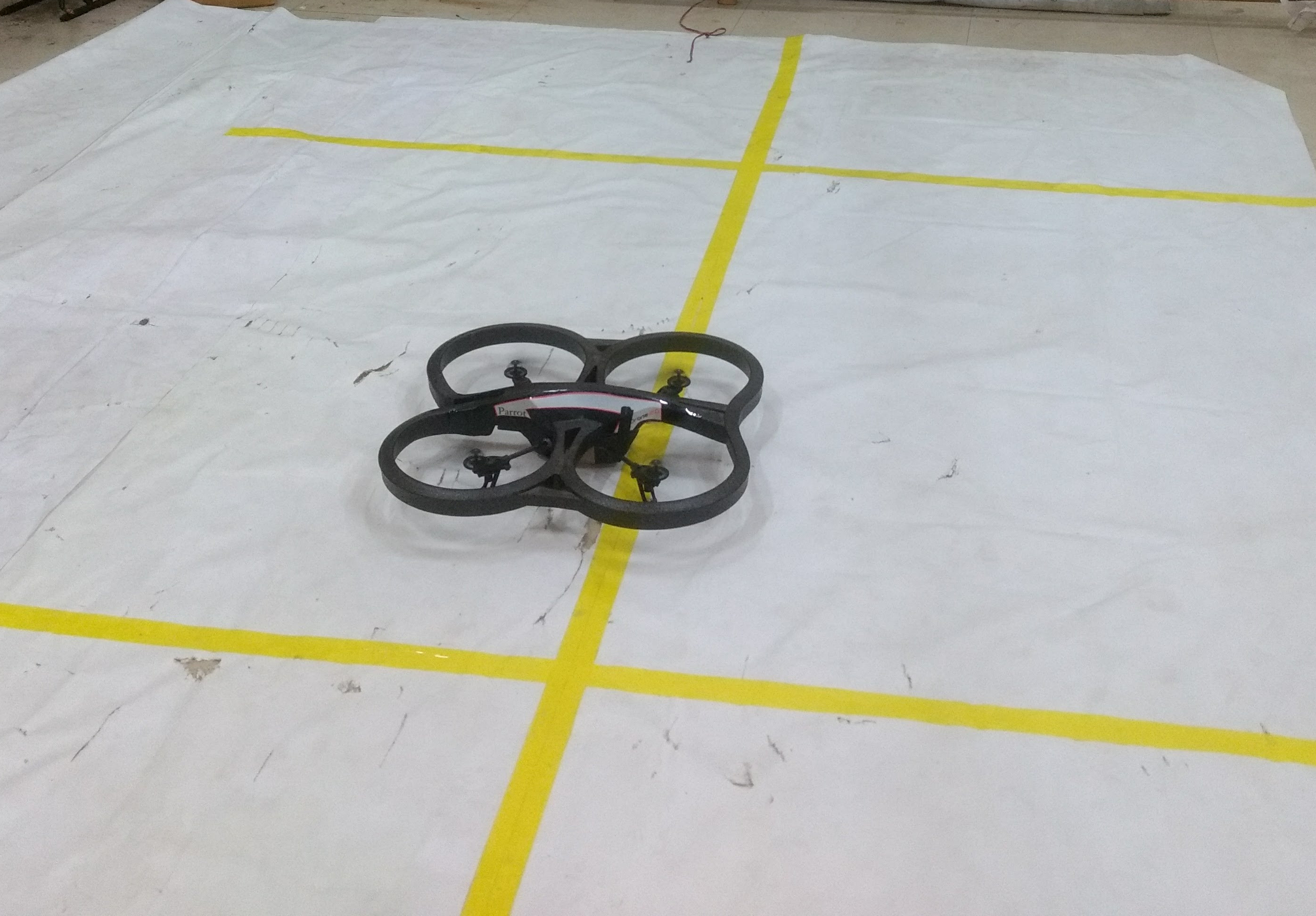}}}
  \caption{Experimental setup for AR drone's PID tuning on grid lines using roll, pitch and yaw as control variables}
\end{figure}

\begin{table}[ht!]
\caption{Performance evaluation of our method on different hardware.}
\label{table_example}
\begin{center}
\begin{tabular}{| c | c | c |}
\hline
\textbf{Method} & \textbf{On-board} (fps) & \textbf{Off-board} (fps)\\
\hline
Centroid & 4.23 & 15.10\\

Skeletonization & 8.93 & 20.0\\
\hline
\end{tabular}
\end{center}
\end{table}

\begin{table}[ht]
\caption{Performance evaluation of different method on a dataset of images of size 1920$\times$1080.}
\begin{center}
\begin{tabular}{| c | c | c | c | c | c | c |}
\hline
\multirow{2}{*}{\bfseries Method} & 
\multicolumn{2}{| c |}{\bfseries Simulation} &  
\multicolumn{2}{| c |}{\bfseries Real-world} \\ \cline{2-5} 
 & $\Delta r$ (px) & $\Delta \alpha$ ($\degree$) & $\Delta r$ (px) & $\Delta \alpha$ (\degree) \\ 
\hline
Naive approach & 56 & 3 & 78 & 12 \\
Centroid & 10 & 3 & 18 & 5 \\
Skeletonization & 5 & 4 & 15 & 10 \\
Combined & 6 & 3 & 16 & 7 \\
\hline
\end{tabular}
\end{center}
\end{table}

\section{CONCLUSIONS}

The proposed pipeline demonstrates promising result in both simulation world and real world scenario. Both of our line detection algorithms get away the limitations of pure hough line transform. Depending upon the hardware available, we show the capability of the method to outperform classical techniques. Our clustering algorithm on rho, theta of multiple lines effectively gives the closest node to which the intermediate target should be set to. At any point in time, the offset from this set-position is reduced iteratively by multiple PID loops. This simple yet effective localization method gives a competitive performance regarding speed as compared to the application of heavy SLAM algorithms on same computing power. Though we cannot expect comparable accuracy as we discretized the locations of the drone to nodes co-ordinates, the algorithms achieve decent accuracy in terms of node coordinates as the application area doesn't require per position localization in a well-defined environment. The proposed navigation strategy makes the drone capable of planning a sub-optimal route to attain global target/setpoint. We finally end our conclusion with the restatement that monocular camera-based vision techniques when designed appropriately can perform significantly well as compared to price one needs to pay for using full SLAM-based solutions.



\section*{ACKNOWLEDGMENT}

We would like to thank the institute authorities at Indian Institute of Technology Kharagpur (session 2017-18) and IIT Gymkhana body to avail us the funds as a support for the presented work. Also, a special thanks to members of Aerial Robotics Kharagpur group for providing us their working space and insights time to time for the betterment of the project.


\addtolength{\textheight}{-15cm}   

\bibliographystyle{IEEEtran}
\bibliography{IEEEexample}

\end{document}